\documentclass[journal]{IEEEtran}
\usepackage{stfloats}
\usepackage{amsmath,epsfig}
\usepackage{amsmath,graphicx}
\usepackage{subfigure}
\usepackage{overpic}
\usepackage{makecell}
\usepackage{tabularx}
\usepackage[noadjust]{cite}
\usepackage{multirow}
\usepackage{array}
\usepackage{adjustbox}
\usepackage{supertabular}
\usepackage{times}
\usepackage{rotating}
\usepackage{graphicx}
\usepackage{amsmath}
\usepackage{amssymb}
\usepackage{subfigure}
\usepackage{graphicx}
\usepackage{multirow}
\usepackage{color}
\usepackage{array}
\usepackage{hyperref}
\usepackage{flushend}
\ifCLASSINFOpdf
\else
\fi
\begin{document}
\title{Super-Resolving Face Image by Facial Parsing Information}
%Michael~Shell,~\IEEEmembership{Member,~IEEE,}
        % John~Doe,~\IEEEmembership{Fellow,~OSA,}
        % and~Jane~Doe,~\IEEEmembership{Life~Fellow,~IEEE}

\author{Chenyang Wang, Junjun Jiang, Zhiwei Zhong, Deming Zhai, and Xianming Liu 
%\thanks{This work is supported by the National Science Foundation of China under Grants 61922027, 61932022, and 61971165.}
  \thanks{Corresponding author: Junjun Jiang}
\IEEEcompsocitemizethanks{
\IEEEcompsocthanksitem C. Wang, J. Jiang, Z. Zhong, D. Zhai and X. Liu are with the School of Computer Science and Technology, Harbin Institute of Technology, Harbin 150001, China E-mail: \{wangchy02, jiangjunjun, zhwzhong, zhaideming, csxm\}@hit.edu.cn.
}

 }%
% \thanks{M. Shell was with the Department
% of Electrical and Computer Engineering, Georgia Institute of Technology, Atlanta,
% GA, 30332 USA e-mail: (see http://www.michaelshell.org/contact.html).}% <-this % stops a space
% \thanks{J. Doe and J. Doe are with Anonymous University.}% <-this % stops a space
% \thanks{Manuscript received April 19, 2005; revised August 26, 2015.}}


\markboth{Journal of \LaTeX\ Class Files,~Vol.~14, No.~8, August~2015}%
{Shell \MakeLowercase{\textit{et al.}}: Bare Demo of IEEEtran.cls for IEEE Journals}



\maketitle

\begin{abstract}
Face super-resolution is a technology that transforms a low-resolution face image into the corresponding high-resolution one. In this paper, we build a novel parsing map guided face super-resolution network which extracts the face prior (\textit{i.e.}, parsing map) directly from low-resolution face image for the following utilization. To exploit the extracted prior fully, a parsing map attention fusion block is carefully designed, which can not only effectively explore the information of parsing map, but also combines powerful attention mechanism. Moreover, in light of that high-resolution features contain more precise spatial information while low-resolution features provide strong contextual information, we hope to maintain and utilize these complementary information. To achieve this goal, we develop a multi-scale refine block to maintain spatial and contextual information and take advantage of multi-scale features to refine the feature representations. Experimental results demonstrate that our method outperforms the state-of-the-arts in terms of quantitative metrics and visual quality. The source codes will be available at \url{https://github.com/wcy-cs/FishFSRNet}.
\end{abstract}

\begin{IEEEkeywords}
Face hallucination, face super-resolution, facial prior, multi-scale, parsing map
\end{IEEEkeywords}

% \fi
\IEEEpeerreviewmaketitle

\section{INTRODUCTION}
\label{sec:intro}

%I wish you the best of success.

 

% \subsection{Subsection Heading Here}
% Subsection text here.

%\IEEEpubidadjcol

\IEEEPARstart{F}{ace} super-resolution (FSR), also known as face hallucination, aims to recover a high-resolution (HR) face from a low-resolution (LR) face. In real-world scenarios, due to the limitations of low-cost cameras and the influence of imaging conditions, the captured face images are always in low resolution, which not only provides the user poor visual perception, but also has adverse effects on face-related tasks, such as face attribute analysis~\cite{zheng2020survey}, face detection, face recognition \cite{ATfacegan}, \textit{etc}. Thus, FSR has a wide range of application and drawn increasingly attention in recent years.

In the literature, Baker and Kanada~\cite{C1} propose face hallucination for the first time and recover HR face by searching local features from training set. After that, more and more representative methods are developed. In the early stage, researchers mainly design shallow learning-based methods by exploring the energy of local linear embedding~\cite{Chang}, eigentransformation~\cite{Wang}, principal component analysis~\cite{SRpca} and others. Due to the insufficient representation capability of these methods, they are hard to generate an outstanding HR face image, especially when the upscale factor is large, such as $\times$8, $\times$16.%, or the observed LR face is non-frontal and misaligned to the training samples.

Recently, motivated by the success of deep learning, a variety of deep learning-based FSR methods are proposed and have achieved great breakthrough. In the early stage, Zhou \textit{et al}.~\cite{BCCNN} develop the first convolution neural network-based FSR method. Then, Yu \textit{et al}.~\cite{URDGN} develop URDGN, which is a generative adversarial network-based FSR method. In~\cite{Attention-FH,Attention-FHone}, inspired by human perception, the researchers combine deep learning and reinforcement learning to recover HR progressively. In ~\cite{WaveletSRnet}, WSRNet is built to recover faces in wavelet domain. The work of \cite{HiFaceGAN} develops a suppression module to encode semantic information for FSR. Chen \textit{et al}. \cite{SPARNet} build a facial attention for learning local details. Recently, SISN \cite{SISN} designs an internal-feature split attention for capturing internal correlation-ship among intermediate features and then improving the quality of face images.

Human face is a highly structural object, and the face image has its inherited characteristics, such as facial heatmap, facial parsing map, and others, which can improve FSR performance. Then, many methods propose to use the priors to guide the reconstruction process. Zhu \textit{et al}.~\cite{CBN} design a two-branch network for FSR and face correspondence field estimation, respectively. Super-FAN~\cite{super-fan} first recovers a super-resolved face and then estimates the heatmaps of super-resolved faces. To utilize the heatmap, Super-FAN forces the heatmap of super-resolved face to be close to the one of HR face, which is called heatmap loss. However, heatmap loss only works in the training phase, but does not participate in the inference phase of FSR. To solve this problem, some methods have been proposed to insert the heatmap estimation into the super-resolution network. The approach of \cite{FSRFCH} first generates the intermediate features and then estimates the heatmaps from the intermediate features, and then fuses the estimated heatmaps and intermediate features for the following FSR procedure. FSRNet~\cite{FSRNet} recovers a coarse super-resolved face, then estimates face-specific information which is used to assist the next fine reconstruction. Ma \textit{et al}.~\cite{DIC} develop a deep FSR model with iterative collaboration (DIC). It iteratively performs FSR and prior estimation to facilitate two tasks each other. 

 \begin{figure}[t]%[htb]

% \begin{minipage}[b]{1.0\linewidth}
  \centering
  \centerline{\includegraphics[width=\linewidth]{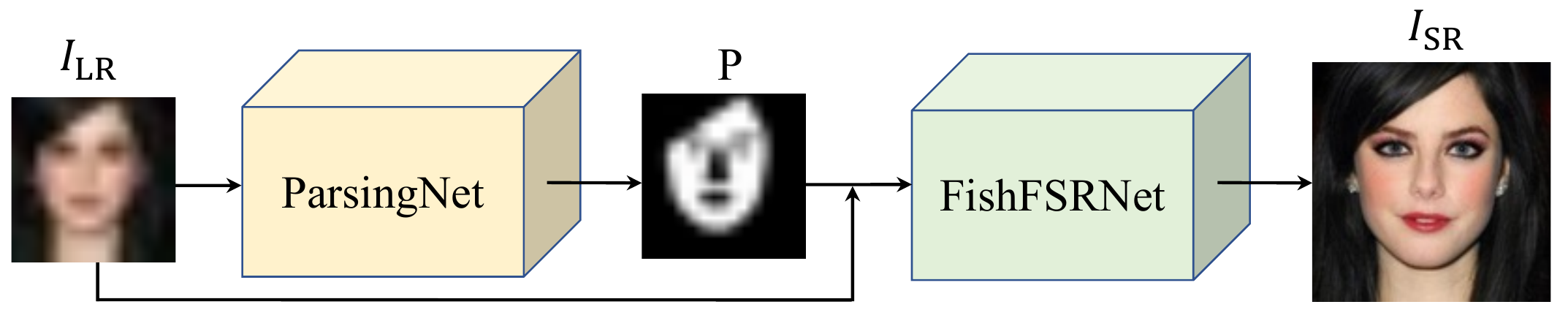}}
%  \vspace{2.0cm}
%  \centerline{(a) Result 1}\medskip
  \caption{The overall framework of our method.}
  \label{net}
% \end{minipage}

\end{figure}

 \begin{figure*}[t]

%\begin{minipage}[b]{1.0\linewidth}
  \centering
  \centerline{\includegraphics[width=\linewidth]{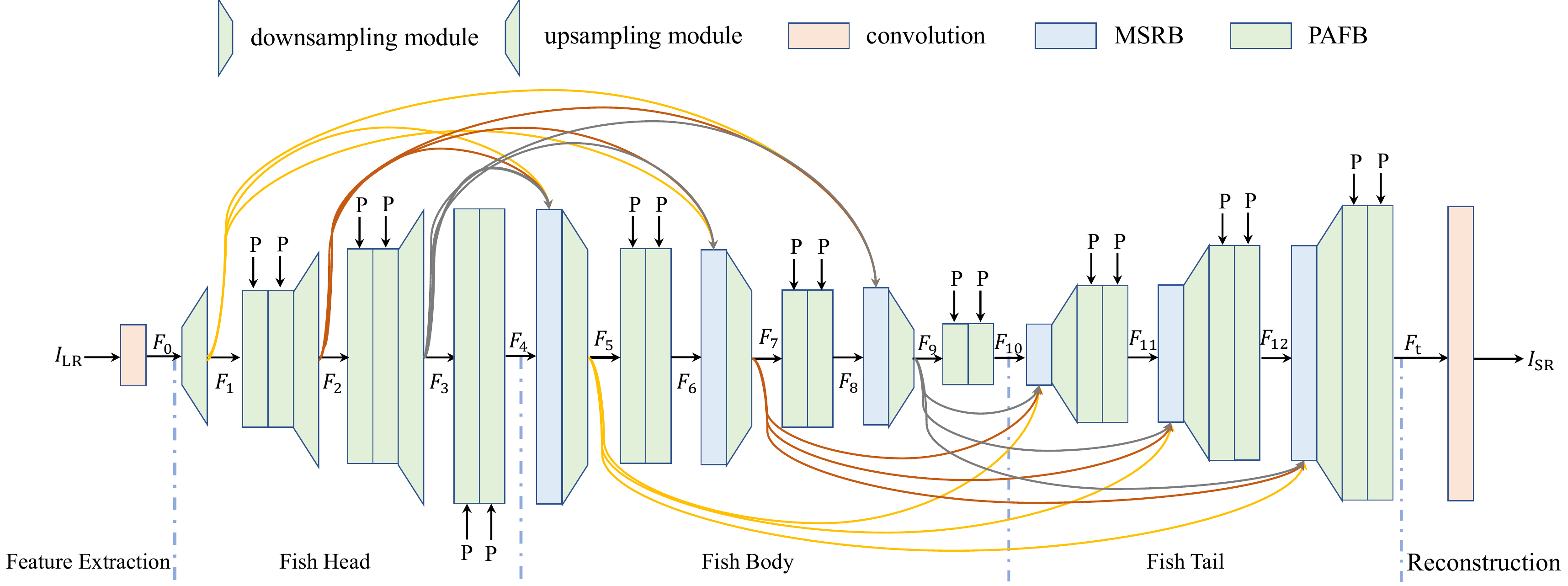}}
%  \vspace{2.0cm}
%  \centerline{(a) Result 1}\medskip
  \caption{The network architecture design of our proposed FishFSRNet. The FishFSRNet consists of five parts: feature extraction layer, fish head, fish body, fish tail and reconstruction layer.}

%\end{minipage}
  \label{FishFSRNet}
\end{figure*}

Although these prior-guided FSR methods have achieved impressive performance, the following concerns still need to be carefully considered: 
i) On the one hand, methods using prior-based loss like \cite{super-fan,PFSRNet} only apply loss in the training phase and the prior information does not participate in the inference phase, which cannot fully capture the potential of the prior information. On the other hand, the prior knowledge derived from the intermediate results is directly affected by the quality of intermediate results, which is usually limited, leading to poor and even wrong prior knowledge learned. ii) Existing deep FSR models always ignore the fusion and utilization of multi-scale (or multi-resolution) features. HR features contain spatially-precise information while LR features have rich contextual information. Due to their complementarity, they should be all kept and used for refining other features. However, most methods are not able to take full advantage of this point.

Considering the shortcomings mentioned-above, we propose a new parsing map attention fusion network for face hallucination. Our method is a two-step network, which is shown in Fig.~\ref{net}. To address the first problem, we develop the \textit{ParsingNet} to extract parsing map directly from LR instead of the intermediate results in \cite{FSRNet,FSRFCH,DIC}. This change can avoid the accumulation of error influenced by inaccurate intermediate results. Then, we recover the final super-resolved face images with the extracted parsing map and LR input by a fish-shape network, \textit{FishFSRNet}. For capturing and utilizing multi-scale features fully, we develop a multi-scale refine block (MSRB) that reserves multi-scale features from the previous layers and uses the reserved features to refine the current features. In addition, we build a parsing map attention fusion block (PAFB) which combines the effectiveness of attention mechanism and parsing map for recovering the important details and face contours.

The main contributions of the proposed method are summarized as follows:
\begin{itemize}
\item We propose a two-step deep face hallucination method which first builds ParsingNet to directly estimate facial parsing map from LR to avoid the wrong information caused by the intermediate results, and then feeds the parsing map and LR into FishFSRNet to recover HR face.
\item We develop a multi-scale refine block (MSRB) to reserve previous features at different scales and refine the current features. In this way, the information of all resolution of face images can be well captured and utilized. 
\item We develop a parsing map attention fusion block (PAFB) that can not only exploit the channel and spatial correlation of features but also make full use of parsing map.
\item Our method achieves the state-of-the-art performance in multiple upscale factors (\textit{e.g.}, $\times$4, $\times$8, $\times$16). Extensive experiments demonstrate the superiority of our method over existing deep learning-based FSR methods in terms of visual and quantitative results.
\end{itemize}

This work is an extension of our previous work of \cite{our}. The essential improvements over our previous work include: i) We have provided a more comprehensive review about related work; ii) We have developed a much more powerful network architecture by exploiting rich features. Considering that features in different resolutions have different information, we have improved the network architecture and designed a multi-scale refine block to maintain and utilize multi-resolution features; iii) We have developed a novel parsing map attention fusion block which can combine the potential of parsing map and attention mechanism and make full use of them; iv) We have conducted more comprehensive experiments and ablation studies to verify the effectiveness of our method.

\section{RELATED WORK}
In the following, we will introduce the related work of FSR, including shallow learning and deep learning-based methods.

\subsection{Shallow Learning-based FSR}
Face hallucination is first proposed by Baker and Kanade~\cite{C1} in 2000, and they enhance resolution by searching a similar structure from trainset. Since then, face hallucination has attracted increasingly attention, and a series of works have been proposed. Liu \textit{et al}.~\cite{Ce} design a two-step FSR method that first uses global linear model to recover a coarse result, and then designs a patch-based nonparametric Markov network for compensating high-frequency details. Inspired by locally linear embedding, Chang \textit{et al}.~\cite{Chang} recover a high-quality face with neighbor embedding. Then, Wang \textit{et al}.~\cite{Wang} take advantage of eigentransformation to recover high quality faces. The work of ~\cite{SRpca} utilizes kernel principal component analysis prior model to extract valuable information for boosting face reconstruction. Except that, there are still many methods utilizing convex optimization~\cite{convex}, canonical correlation analysis~\cite{Huang}, Bayesian approach~\cite{bay}, local structure prior~\cite{7547257}, kernel regression~\cite{8638960}, \textit{etc}. to recover a high quality face. However, these methods fail to restore high-quality faces when the upsampling factor is large (\textit{i.e.}, $\times$8, $\times$16).

\subsection{Deep Learning-based FSR}

\subsubsection{General FSR Methods}
Recently, with the booming development of deep learning in computer vision \cite{8918046,9394762}, deep learning-based face hallucination methods have achieved great breakthroughs \cite{jiang2021deep}. In the early stage, scholars mainly design efficient network structures for FSR without considering face specificity and these methods are called general FSR methods. Zhou \textit{et al}.~\cite{BCCNN} make the first attempt to super-resolve face images with convolution neural network. In detail, URDGN~\cite{URDGN} constructs network based on generative adversarial network. Considering that LR can be unaligned, TDN~\cite{TDN,MTDN} inserts spatial transformer networks~\cite{STN} into the network to perform the alignment. To enhance robustness of the model, Yu \textit{et al}.~\cite{TDAE} build transformative discriminative autoencoders for noisy and unaligned face repair. Considering that the LR face images may be non-frontal, the work of \cite{FH1} proposes to jointly perform FSR and face frontalization. \cite{ATfacegan} and \cite{9762752} develop effective networks to recover face images degraded by atmospheric turbulence. After that, inspired by that human perception process in which human starts from whole images and then explore a sequence of regions with an attention shifting mechanism, Cao \textit{et al}.~\cite{Attention-FH,Attention-FHone} combine reinforcement learning to recover faces gradually. Jiang \textit{et al}. \cite{jjj} ﬁrst recovers a smooth noise-free intermediate result, then takes advantage of high-quality trainset and facial landmarks to compensate high-frequency details for the smooth intermediate results. Different from the aforementioned methods that recover faces in image domain, Huang \textit{et al}.~\cite{WaveletSRnet,WaveletSRNet1} find that wavelet transformation can depict the textural and contextual information of the images, thus they super-resolve faces in the wavelet domain and propose WSRNet. Jiang \textit{et al}. \cite{ATMFN} utilizes ensemble learning to recover HR faces. Recently, the work in~\cite{SPGAN} develops a supervised pixel-wise GAN to improve FSR performance. Dou \textit{et al}.~\cite{PCA-SRGAN} apply PCA to recover face images progressively. \cite{E-ComSupResNet} develops a compact network and introduces global learning strategy for face hallucination. Chen \textit{et al}.~\cite{SPARNet} design a facial attention to recover local details well while Lu \textit{et al}.~\cite{SISN} develop internal-feature split feature for capturing facial semantic information. 

\subsubsection{Prior-guided FSR Methods}
Instead of focusing on network architecture design, prior-guided FSR methods always utilize facial prior (\textit{i.e.}, geometric prior, reference prior, dictionary prior, codebook prior, generative prior, \textit{etc}.) to improve FSR performance. We introduce them in detail as follows.

\textbf{Geometric prior-based methods: }Since face image is a highly structured object and has face-specific geometric prior (including facial heatmaps, facial landmarks, facial parsing map, \textit{etc}.) which is helpful for facial detail reconstruction, geometric prior-based FSR methods are proposed. At first, CBN~\cite{CBN} recovers face images and estimates face-specific information (dense correspondence field) progressively. However, CBN needs complex and extensive preprocess. Later on, Super-FAN~\cite{super-fan} constrains the facial heatmaps of super-resolved results should be close to the ones of HR faces. FSRNet~\cite{FSRNet} builds a coarse FSR network to produce a coarse super-resolved result which is used to estimate facial prior information and extract features, finally the extracted prior and features are fed into fine FSR network, generating the final results. Yu \textit{et al}.~\cite{FSRFCH} embed facial heatmap estimation into FSR network. Ma \textit{et al}.~\cite{DIC} iteratively perform FSR and estimate heatmaps which are used for performing FSR in the next iteration (DIC). In addition, DIC develops an attention fusion module to utilize facial prior. Different from these methods, JASRNet~\cite{JASRNet} has a share encoder for FSR and facial prior estimation to extract features with maximum information. After that, Li \textit{et al}.~\cite{ATENet1} design a structure enhancement network to estimate and utilize facial boundaries. HCRF \cite{hcrf} combines random forest to recover different facial components effectively. Yu \textit{et al}.~\cite{9321495} build a semantic-driven face hallucination network and develop an improved residual block to combine facial prior. Wang \textit{et al}. \cite{wang2022propagating} design a prior distillation framework to recover face images. Instead of using 2D prior, Hu \textit{et al.} \cite{FSRG3DFP} focus on the utilization of 3D prior. Among these works, FSRNet \cite{FSRNet} is the most related to our method. Here we analyze the difference between them. Firstly, FSRNet extracts prior from coarse FSR result and utilizes it by concatenation while our method extracts prior from LR face, designs PAFB to finely fuse prior at multiple scale and builds MSRB to utilize complementary multi-scale information.

\textbf{Reference prior-based methods: }In addition to explore geometric prior, some researchers focus on reference prior extracted from high-quality reference face image. Originally, researchers use high-quality face images with the same identity with LR face as reference. The works of \cite{GFRNet,GWAINet} first align the reference with LR and then extract identity-aware features from reference to improve face image quality. In case of having many reference images with the same identity, ASFFNet \cite{ASFFNet} designs guidance selection module to select the best reference which has the most similar pose, and then develops adaptive feature fusion block to combine reference information.

\textbf{Dictionary prior-based methods: }Considering that different people may have similar facial components, facial component dictionary constructed from a high-quality face dataset is used to guide FSR. Li \textit{et al.} \cite{DFDNet} first extract feature from the entire dataset and then crop the facial component and then clustering K classes to form dictionary. After that, it designs dictionary transfer module to transfer high-frequency features from dictionary to LR features.

\textbf{Codebook prior-based methods: } Apart from facial component dictionary which can represent facial structure explicitly, discrete codebook prior presenting facial information implicitly is also utilized \cite{codeformer,restoreformer,vqfr,9878755}. To be specific, the codebook is constructed by VQVAE \cite{vavae} with high-quality face dataset. Zhao \textit{et al}. \cite{9878755} map LR face into code space and then find the most similar code in code book as reference code, then replace the LR code with reference code to reconstruct high-quality face by reconstruction branch while Gu \textit{et al.} \cite{vqfr} introduce an additional texture branch to capture texture information. Instead of replacing the LR code with reference code, Wang \textit{et al}. \cite{restoreformer} use multi-head self-attention to fuse LR code and reference code. However, codebook is constructed by high-quality face image while LR code is from LR face image, directly finding the most similar code between them is unreasonable. In view of this, the work of \cite{codeformer} designs a transformer module to predict code sequence.

\textbf{Generative prior-based methods: }Rather than learning a codebook, the works of \cite{yanggan,zhublind,wangtowards,glean,pulse} pretrain a generative model with a high-quality face dataset and use the generator as generative prior. Specially, PULSE \cite{pulse} directly uses the generator to generate SR result while \cite{yanggan} builds an encoder to obtain latent code from LR and feeds the latent code to the generator to generate SR result. Different from them, method in \cite{wangtowards,glean} only use the synthesis network of the generator to assist FSR. Zhu \textit{et al}. \cite{zhublind} combine generative prior and geometric prior to perform blind face restoration.

\section{METHODS}
\label{sec:format}
\begin{figure}%[htb]
  \centering
  \centerline{\includegraphics[width=0.8\linewidth]{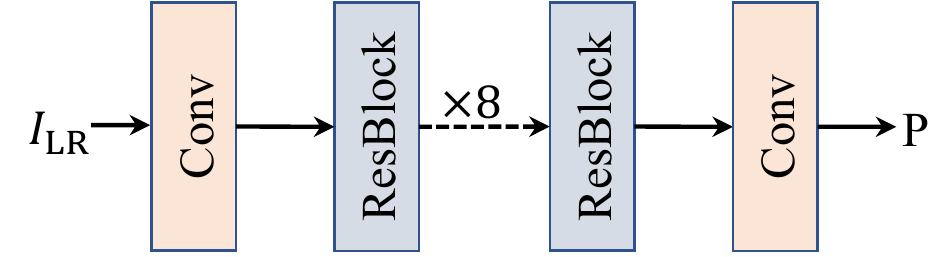}}
  \caption{The architecture of our proposed ParsingNet.}
\label{parsingnet}
\end{figure}
Face super-resolution (FSR) aims to recover a high-resolution (HR) face from the corresponding low-resolution (LR) one, in which face-specific prior knowledge is expected to boost the performance. To achieve this goal, we develop a two-stage network as illustrated in Fig.~\ref{net}. Overall, our method first estimates face-specific information \textit{i.e.}, a parsing map, from LR by our ParsingNet, and then feeds the LR and prior into the super-resolution network (FishFSRNet) to recover the HR one. In FishFSRNet, we develop a multi-scale refine block (MSRB) which reserves and utilizes multi-scale features to refines the current features. Moreover, we build a novel parsing map attention fusion block (PAFB) to explore and combine the potential of parsing map and attention mechanism. 

\subsection{ParsingNet}
Different from general images which have various objects and landscapes, one aligned face image always has a face in the center and has face-specific information. Thus, FSR methods tend to explore the utilization of face-specific information, including facial landmarks, facial parsing map, facial heatmaps and others. In this paper, we explore the potential of facial parsing map in FSR. In this section, we propose a parsing map prediction network named ParsingNet, which predicts a facial parsing map directly from LR. Considering that the LR face images are low-quality and extracting accurate facial parsing map from LR is very difficult, we simplify the task and decrease the difficulty of the task. Specifically, a simple parsing map which is a 0-1 matrix where the skin is 1 and the facial components and others are 0, is used as the facial prior. Then we feed the LR face images into the ParsingNet, and obtain the parsing maps, which can be formulated as 
\begin{equation}
    P = f_{\text{ParsingNet}}(I_{\text{LR}}),
\end{equation}
where $f_{\text{ParsingNet}}$ is the function of our ParsingNet, and $P$ is the parsing map estimated by the ParsingNet. As shown in Fig. \ref{parsingnet}, our ParsingNet is comprised of two convolutional layers and eight Resblocks \cite{Resblock}. To force our ParsingNet to estimate accurate facial parsing map, we design a parsing map loss, which can be defined as
 \begin{equation}
     \mathcal{L}_{\text{ParsingNet}}=\left\|P-Y\right\|_{\text{1}},
 \end{equation}
 where $Y$ is the ground truth of facial parsing map.

\subsection{FishFSRNet}

After introducing ParsingNet, we come to the super-resolution network FishFSRNet. HR features contain spatially-precise information while LR features have rich contextual information. Due to their complementarity, they should be all kept and used for refining other features. In view of this, we aim to design a network that can extract and utilize multi-scale features. To achieve this goal, we develop our FishFSRNet which consists of a feature extraction layer, fish head, fish body, fish tail and a reconstruction layer, as shown in Fig.~\ref{FishFSRNet}. In our FishFSRNet, the LR faces are first progressively upsampled, and then downsampled and finally upsampled to generate the final super-resolved face images. In this pattern, our FishFSRNet can obtain multi-scale feature at different layers and utilize these features for boosting FSR. 

Specifically, given the LR face image $I_{\text{LR}}$, the FishFSRNet extracts features from it by the feature extraction layer which is a $3\times3$ convolutional layer,
\begin{equation}
    F_{\text{0}}=f_{\text{Feature Extraction}}(I_{\text{LR}}),
\end{equation}
where $f_{\text{Feature Extraction}}$ is the function of the feature extraction layer, and $F_{\text{0}}$ is output feature. Then, extracted feature is fed into the fish head. To extract multi-scale features and increase the receptive field, our fish head upsamples the $F_{\text{0}}$ three times. Moreover, to promote the utilization of the multi-scale features, the result of each upsampling is preserved and passed to the following layers. Thus, the process of fish head can be formulated as:
\begin{equation}
    F_{\text{1}}, F_{{\text{2}}}, F_{\text{3}}, F_{\text{4}}=f_{\text{Fish Head}}(F_{\text{0}}, P),
\end{equation}
where $f_{\text{Fish Head}}$ is the function of fish head, and $F_{\text{1}}, F_{\text{2}}, F_{\text{3}}$ are the multi-scale features that are preserved, and $F_{4}$ is the final output of fish head.
\begin{itemize}
    \item To be specific, $F_{\text{0}}$ is first upsampled to obtain $F_{\text{1}}$ which is preserved for the following layers,
\begin{equation}
    F_{\text{1}}=f_{\text{Up}}(F_{\text{0}}),
\end{equation}
where $f_{\text{Up}}$ denotes an upsampling module which is comprised of a 3$\times$3 convolutional layer and a $\times$2 pixelshuffle~\cite{pixelshuffle} layer. Then, $F_{\text{1}}$ and parsing map $P$ are passed through two cascaded PAFBs to explore facial prior information, and the results are upsampled again to acquire the feature in $\times$4 resolution,
\begin{equation}
    F_{\text{2}}=f_{\text{Up}}(f_{\text{PAFBs}}(F_{\text{1}}, P)),
\end{equation}
where $F_{\text{2}}$ is the output by $\times$4 upsampling and it is preserved for the following utilization, and $f_{\text{PAFBs}}$ corresponds to two cascaded PAFBs. Then, with the parsing map, $F_{\text{2}}$ is processed and upsampled again,
\begin{equation}
    F_{\text{3}}=f_{\text{Up}}(f_{\text{PAFBs}}(F_{\text{2}}, P)),
\end{equation}
where $F_{\text{3}}$ is the output upscaled by $\times$8. Then, the last PAFBs in the fish head are applied,
\begin{equation}
    F_{\text{4}}=f_{\text{PAFBs}}(F_{\text{3}}, P),
\end{equation}
where $F_{\text{4}}$ is the final output of the fish head. 
\end{itemize}

Then, come to the fish body. In contrast to the fish head, fish body downsamples the features three times to generate multi-scale features, and inserts a multi-scale refine block (MSRB) before the first downsampling module, and every downsampling module is followed by two PAFBs. The MSRB aims to utilize multi-scale features ($F_{\text{1}}, F_{{\text{2}}}, F_{\text{3}}$) passed from the previous fish head for refining the current features. The process of the fish body can be formulated as:
\begin{equation}
F_{\text{5}}, F_{\text{7}}, F_{\text{9}}, F_{\text{10}} = f_{\text{Fish Body}}(F_{\text{4}}, F_{\text{1}}, F_{\text{2}}, F_{\text{3}}, P),
\end{equation}
where $f_{\text{Fish Body}}$ is the function of fish body, and $F_{\text{5}}, F_{\text{7}}, F_{\text{9}}$ are the multi-scale features in different resolutions and $F_{10}$ is the final output. The formulation of fish body is introduced in detail in the following. 
\begin{itemize}
    \item Firstly, MSRB in fish body utilizes the previous multi-scale features to refine the current features for exploring the complementary information in high- and low- resolution features, and downsamples the refinement result,
    \begin{equation}
    F_{\text{5}}=f_{\text{Down}}(f_{\text{MSRB}}(F_{\text{4}}, F_{\text{1}}, F_{\text{2}}, F_{\text{3}})),
\end{equation}
where $f_{\text{Down}}$ is a downsampling module (comprised of a 3$\times$3 convolutional layer and a inverse pixelshuffle \cite{inv} layer to perform $\times$2 downsampling), $f_{\text{MSRB}}$ denotes the MSRB module, and $F_{\text{5}}$ is the output feature refined and downsampled by the MSRB and the downsampling module. Then, two cascaded PAFBs are applied to $F_{\text{5}}$,
\begin{equation}
    F_{\text{6}}=f_{\text{PAFBs}}(F_{\text{5}}, P),
\end{equation}
obtaining $F_{\text{6}}$. After that, the MSRB exploits the multi-scale features $F_{\text{1}}, F_{\text{2}}, F_{\text{3}}$ passed from the previous layers to refine the current feature $F_{\text{6}}$ again,
\begin{equation}
    F_{\text{7}}=f_{\text{Down}}(f_{\text{MSRB}}(F_{\text{6}}, F_{\text{1}}, F_{\text{2}}, F_{\text{3}})),
\end{equation}
where $F_{\text{7}}$ is feature refined by MSRB. Similar to $F_{\text{5}}$, it is also processed by two PAFBs to combine facial prior,
\begin{equation}
    F_{\text{8}}=f_{\text{PAFBs}}(F_{\text{7}}, P),
\end{equation}
generating $F_{\text{8}}$. Then, $F_{\text{8}}$ is passed to incorporate with previous multi-scale features and be downsampled,
\begin{equation}
    F_{\text{9}}=f_{\text{Down}}(f_{\text{MSRB}}(F_{\text{8}}, F_{\text{1}}, F_{\text{2}}, F_{\text{3}})),
\end{equation}
where $F_{\text{9}}$ is the refined result. Until now, $F_{\text{5}}, F_{\text{7}}, F_{\text{9}}$ in three different resolutions are obtained and preserved. Finally, $F_{\text{10}}$ is acquired by PAFBs,
\begin{equation}
    F_{\text{10}}=f_{\text{PAFBs}}(F_{\text{9}}, P).
\end{equation}
\end{itemize}
Similar to the fish head, the fish body also preserves the multi-scale features generated by the downsampling modules for the utilization in the following layers. These features are fed into the fish tail by skip connection. After that, with features from the fish body, our fish tail progressively upsamples features, generating features in the same resolution with the HR face. Similar to the fish head, our fish tail is also comprised of three upsampling modules, PAFBs and MSRBs. The utilization of previous features in fish tail is also finished by MSRBs. The process can be defined as
\begin{equation}
F_{\text{t}}=f_{\text{Fish Tail}}(F_{\text{10}}, F_{\text{9}}, F_{\text{7}}, F_{\text{5}}, P),
\end{equation}
where $f_{\text{Fish Tail}}$ denotes our fish tail and $F_{\text{t}}$ is its output. 
\begin{itemize}
    \item In detail, the process of fish tail can be formulated into three sequential MSRB-Down-PAFBs operations, \begin{equation}
    F_{\text{11}}=f_{\text{PAFBs}}(f_{\text{Down}}(f_{\text{MSRB}}(F_{\text{10}}, F_{\text{9}}, F_{\text{7}}, F_{\text{5}})),P),
\end{equation}
\begin{equation}
    F_{\text{12}}=f_{\text{PAFBs}}(f_{\text{Down}}(f_{\text{MSRB}}(F_{\text{11}}, F_{\text{9}}, F_{\text{7}}, F_{\text{5}})),P),
\end{equation}
\begin{equation}
    F_{\text{t}}=f_{\text{PAFBs}}(f_{\text{Down}}(f_{\text{MSRB}}(F_{\text{12}}, F_{\text{9}}, F_{\text{7}}, F_{\text{5}})),P),
\end{equation}
where $F_{\text{11}}$ and $F_{\text{12}}$ are the intermediate features.
\end{itemize}

Finally, we generate the recovered HR face through the reconstruction layer:
\begin{equation}
I_{\text{SR}}=f_{\text{Reconstruction}}(F_{\text{t}}),
\end{equation} 
where $f_{\text{Reconstruction}}$ denotes the function of the reconstruction layer (a 3$\times$3 convolutional layer), and $I_{\text{SR}}$ is recovered face.

To measure the pixel consistency between the super-resolved results and the ground truth, we use a pixel loss that measures the pixel distance, which can be expressed as
   \begin{equation}
     \mathcal{L}_{\text{FishFSRNet}}=\left\|I_{\text{SR}}-I_{\text{HR}}\right\|_{\text{1}},
 \end{equation}
where $I_{\text{HR}}$ is the ground truth. To supervise the model, we choose L1 loss as our loss function. 

\begin{figure}%[htb]
  \centering
  \centerline{\includegraphics[width=\linewidth]{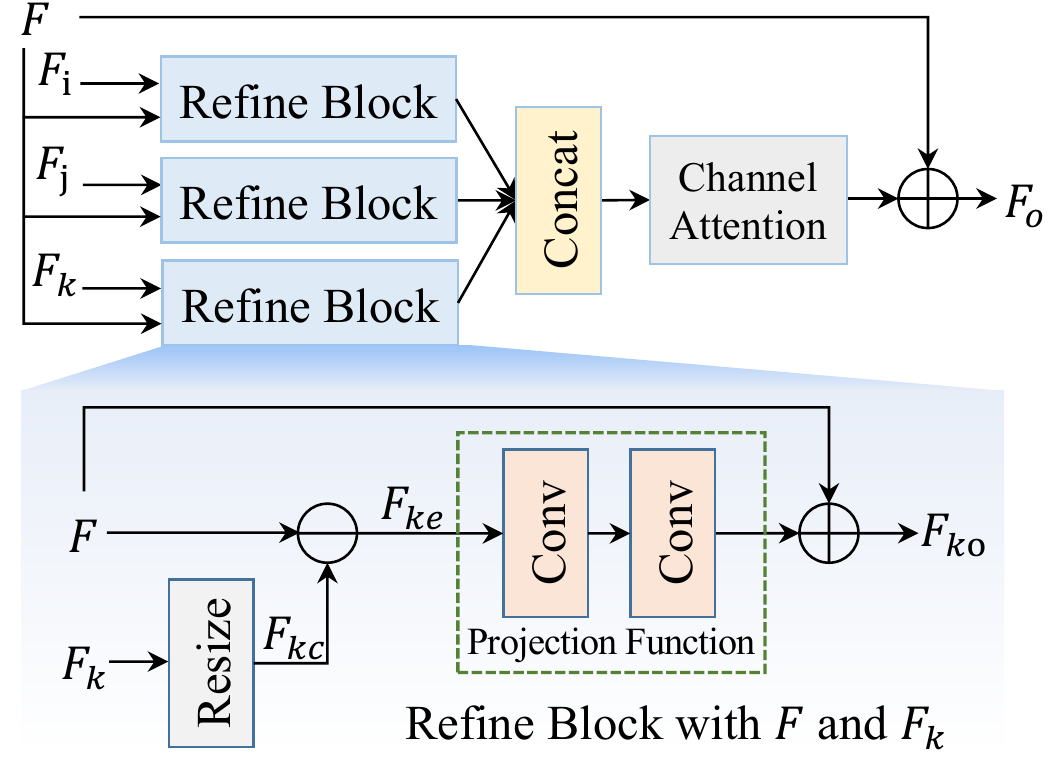}}
  \caption{The architecture design of our proposed multi-scale refine block.}
\label{MSRB}
\end{figure}

\subsection{Multi-scale Refine Block (MSRB)}
HR features tend to contain more spatially-precise information while LR features have rich contextual information. Since the information contained by HR and LR features is complementary, features in different resolutions should be all utilized. In light of this, we first pass all previous features at different scales to every downsampling module in the fish body and every upsampling module in the fish tail. Then, we develop MSRB and insert it before every downsampling module in fish body as well as the upsampling module in fish tail to refine the current features with previous feature of different resolutions. In the following, we will introduce our MSRB and show the architecture of MSRB in Fig.~\ref{MSRB}.

Taking the upscale factor $\times$8 as an example. The input of MSRB includes four parts: the current feature $F$, previous features $F_{\text{i}}, F_{\text{j}}$ and $F_{\text{k}}$. First, we feed $F$ with every previous feature into three refine blocks respectively to refine $F$. After that, we concatenate the refined outputs of refine blocks. Considering that features in LR contain rich contextual information and features in HR provide spatially-precise information, refined features should be treated discriminatively. Thus, we further feed the concatenation result into a channel attention mechanism to capture information along channel dimension. Finally, the sum of $F$ and the output of the channel attention is calculated, which is the final fused and refined feature.

\textbf{Refine block:} Inspired by the back-projection algorithm~\cite{DBPN}, we develop a refine block which is used to utilize the multi-scale features from the previous stage and remedy the missing information of the current feature. In essence, our refine block is designed to enhance and refine the current features with an error feedback mechanism. As shown in Fig.~\ref{MSRB}, for every previous feature (here we take $F_{\text{k}}$ as an example), our refine block first resizes (implemented by the nearest interpolation or a convolution layer) it to generate $F_{\text{kc}}$ that shares the same resolution with $F$. Then, the error feedback mechanism can be implemented.
\begin{itemize}
    \item The first step is to compute the difference $F_{\text{ke}}$ between the current feature $F$ and the previous feature $F_{\text{kc}}$,
    \begin{equation}
        F_{\text{ke}} = F -F_{\text{kc}}.
    \end{equation}
    \item The second step is to update the current feature $F$ with the projected error features:
    \begin{equation}
        F_{\text{ko}} = f_{\text{P}}(F_{\text{ke}})+F,
    \end{equation}
    where $f_{\text{P}}$ is the function of projection implemented by two convolution layers and $F_{\text{ko}}$ is a refined feature.
\end{itemize}

\begin{figure}%[htb]
  \centering
  \centerline{\includegraphics[width=\linewidth]{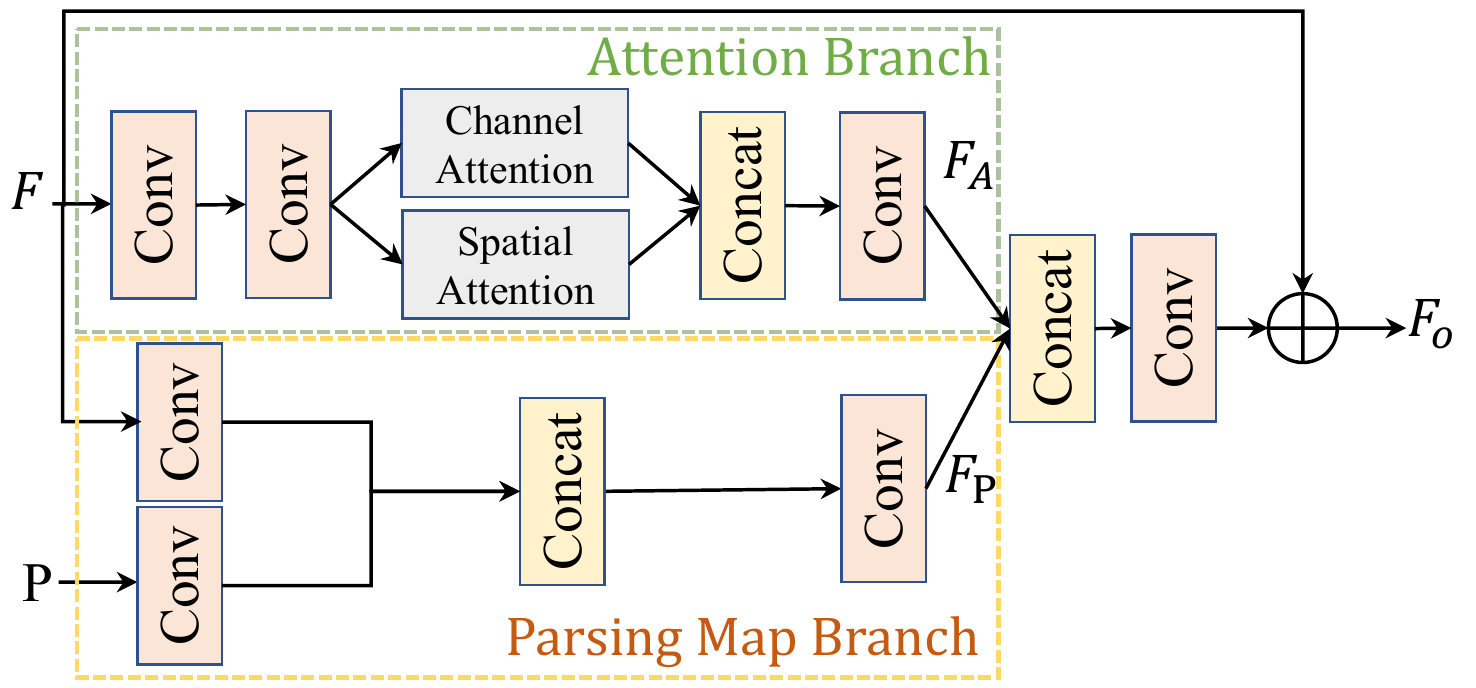}}
  \caption{The architecture of our parsing map attention fusion block.}
\label{PAFB}
\end{figure}
\subsection{Parsing Map Attention Fusion Block (PAFB)}\label{des-pafb}
Recently, attention mechanism, mainly including channel attention mechanism or spatial attention mechanism, is widely used in the image super-resolution field and has achieved a great breakthrough, such as the work in~\cite{RCAN,CSFM}. It is generally accepted that channel and spatial attention mechanisms can extract effective features and boost image super-resolution. We assume that the face image is a special case of general image, thus the technique that can promote general image super-resolution can also benefit face restoration. Depending on this assumption, we introduce channel and spatial attention mechanism into FSR. In addition, face is a high-structured object which has its own face-specific information. Thus, we also take advantage of face-specific prior knowledge (\textit{i.e.}, parsing map) to improve face restoration. Combining face-specific information and an effective attention mechanism, we develop a parsing map attention fusion block (PAFB) which consists of an attention branch and a parsing map branch as shown in Fig.~\ref{PAFB}. Our proposed PAFB can not only explore the information dependency along the spatial and attention dimensions, but also make full use of the parsing map.

\textbf{Attention Branch:} Motivated by the powerful representative ability of attention mechanism in general image super-resolution, we develop an attention branch to explore inter-channel and spatial relationship. To be specific, our attention branch first employs two convolution layers to extract features, and then feed extracted features into parallel channel and spatial attentions to capture inter-channel and spatial dependencies. To effectively fuse the channel and spatial mechanisms, the attention branch directly concatenates the outputs of them and applies a fusion convolution layer to further combine them, generating $F_{\text{A}}$.

\textbf{Parsing Map Branch:} Given that a face has its own face-specific prior knowledge that can boost FSR, we further develop a parsing map branch. First, we interpolate $P$ into the same size with $F$ by nearest interpolation. Since $F$ and parsing map $P$ are in different domains and can provide different information, two convolution layers are applied on them to project them into a similar domain. After that, concatenation followed by a fusion convolution layer is applied on the projected features to simply fuse them, generating $F_{\text{P}}$. %Note that the parsing map is upsampled by nearest interpolation into the same size with F in every PAFB.

Since the attention branch pays more attention to the inter-channel and spatial relationship while the parsing map branch lays stress on the exploration of prior, we further combine them to boost the cooperation between them and promote the performance of FSR. Specifically, concatenation followed by a convolution layer is used to fuse them adaptively. The effectiveness of the two branches is verified in Section~\ref{sec:experiment}.

\subsection{Architecture Novelty of FishFSRNet}
The architecture design novelty of FishFSRNet is mainly reflected in MSRB and PAFB which have their specific characteristics. i) MSRB: In light of that high-resolution features contain more precise spatial information while low-resolution features provide strong contextual information, we hope to maintain and utilize these complementary information. Thus, MSRB is designed. It not only fuses features from previous layers, but also exploits the potential of features in different resolutions. In addition, MSRB has a refinement mechanism to utilize the multi-scale features to refine the current features. ii) PAFB: in PAFB we not only introduce attention mechanism to FSR but also fuse them with the face-specific information (\textit{i.e.,} paring map). Specifically, on the one hand, we use the spatial and channel attention to capture spatial- and channel-wise information implicitly. On the other hand, we embed the face parsing map to explore facial structure information explicitly. By the collaboration between them, the proposed method can achieve a very good performance.

\section{EXPERIMENTS}
\label{sec:experiment}
 \subsection{Datasets and Metrics}
 We conduct extensive experiments on two widely used datasets: CelebA~\cite{celeba} and Helen~\cite{helen}. In our ablation study, we adopt CelebA to verify the effectiveness of every component. In quantitative and qualitative comparisons with state-of-the-art methods, both CelebA and Helen are used.
 
 \textbf{CelebA:} CelebA is a a large-scale face dataset, and contains more than 200,000 faces in large pose diversity and background clutter, providing 5 facial landmarks, 40 attributes and identity information. For CelebA dataset, we use 168,854 faces as training set, and 100 faces as validation set, and 1,000 faces as testing set, following~\cite{DIC}. 
 
 \textbf{Helen:} Helen is a face dataset that contains 2,330 faces under a broad range of appearance variation, providing 194 landmarks. For Helen dataset, we following the setting of~\cite{DIC} and use 2,005 faces for training (wherein 1,955 faces as the training set and 50 faces as validation set), and the remaining 50 faces as the testing set.
 
 Peak Signal-to-Noise Ratio (PSNR) and structural similarity (SSIM)~\cite{ssim} indices are introduced as metrics for evaluation. They are computed on the Y channel of YCbCr space. 

 \subsection{Implementation details}

\textbf{Training setting} Following DIC~\cite{DIC}, we first preprocess the face images. For both datasets, we use 68 landmarks extracted by OpenFace~\cite{openface1,openface2,openface3}. Depending on the face region, we first crop the faces and resize them into 128$\times$128 as ground truth HR faces, and then resize the ground truth into 64$\times$64, 32$\times$32, and 16$\times$16 as corresponding LR faces with upscaling factor $\times$4, $\times$8, and $\times$16 respectively. For facial parsing maps, we adopt pretrained BiSeNet~\cite{BiSeNet} to extract parsing map of different facial components from HR and fuse. For ParsingNet with different scales, we downsamples the extracted parsing map with the corresponding scale to obtain the ground truth parsing map of ParsingNet. 

Our experiments are implemented with the popular toolbox Pytorch~\cite{Pytorch}. We first train the ParsingNet, and then with the estimated parsing map, we train the super-resolution network FishFSRNet. Both ParsingNet and FishFSRNet are optimized by ADAM with $\beta_{1}=0.9$, $\beta_{2}=0.99$ and $\epsilon=1e-8$. The learning rate is set as $1e-4$ in the training phase, and the mini-batch size of the model is set as 8. Note that there are 2, 3, 4 upsampling modules in fish head and fish tail, 2, 3, 4 downsampling modules in fish body for $\times$4, $\times$8, and $\times$16 FSR respectively, and every upsampling or downsampling module is followed by two cascaded PAFBs.
 \begin{figure}[t]
	\centering
    \includegraphics[width=\linewidth]{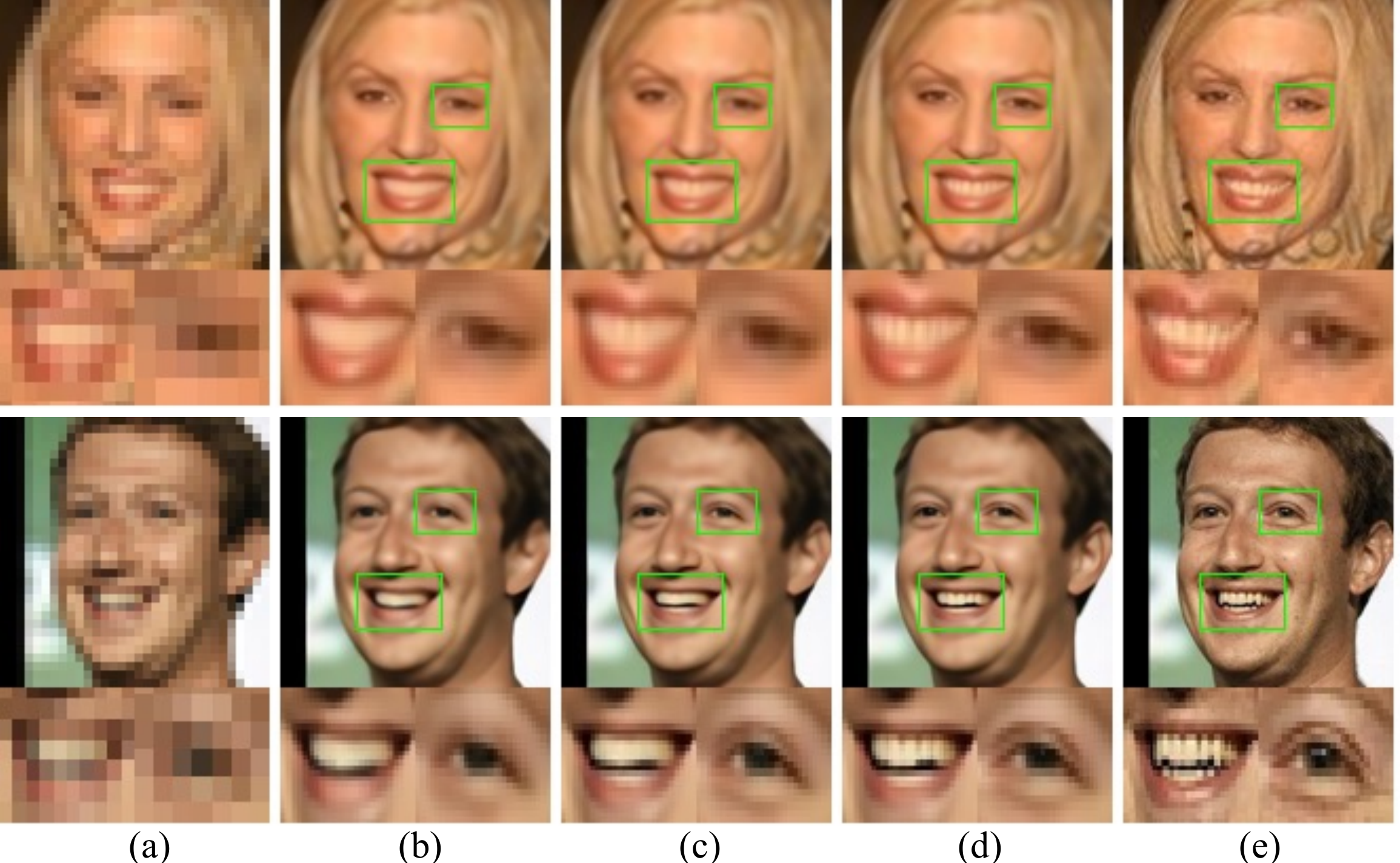}
	\caption{Visual quality comparison of state-of-the-art methods by the scale of $\times$4. Please zoom in to view the differences. (a): LR; (b): DIC~\cite{DIC}; (c): SISN~\cite{SISN}; (d): Ours; (e): HR.}
	\label{fig:upscale_x4}
\end{figure}
      \begin{figure*}[t]
	\centering
    \includegraphics[width=\linewidth]{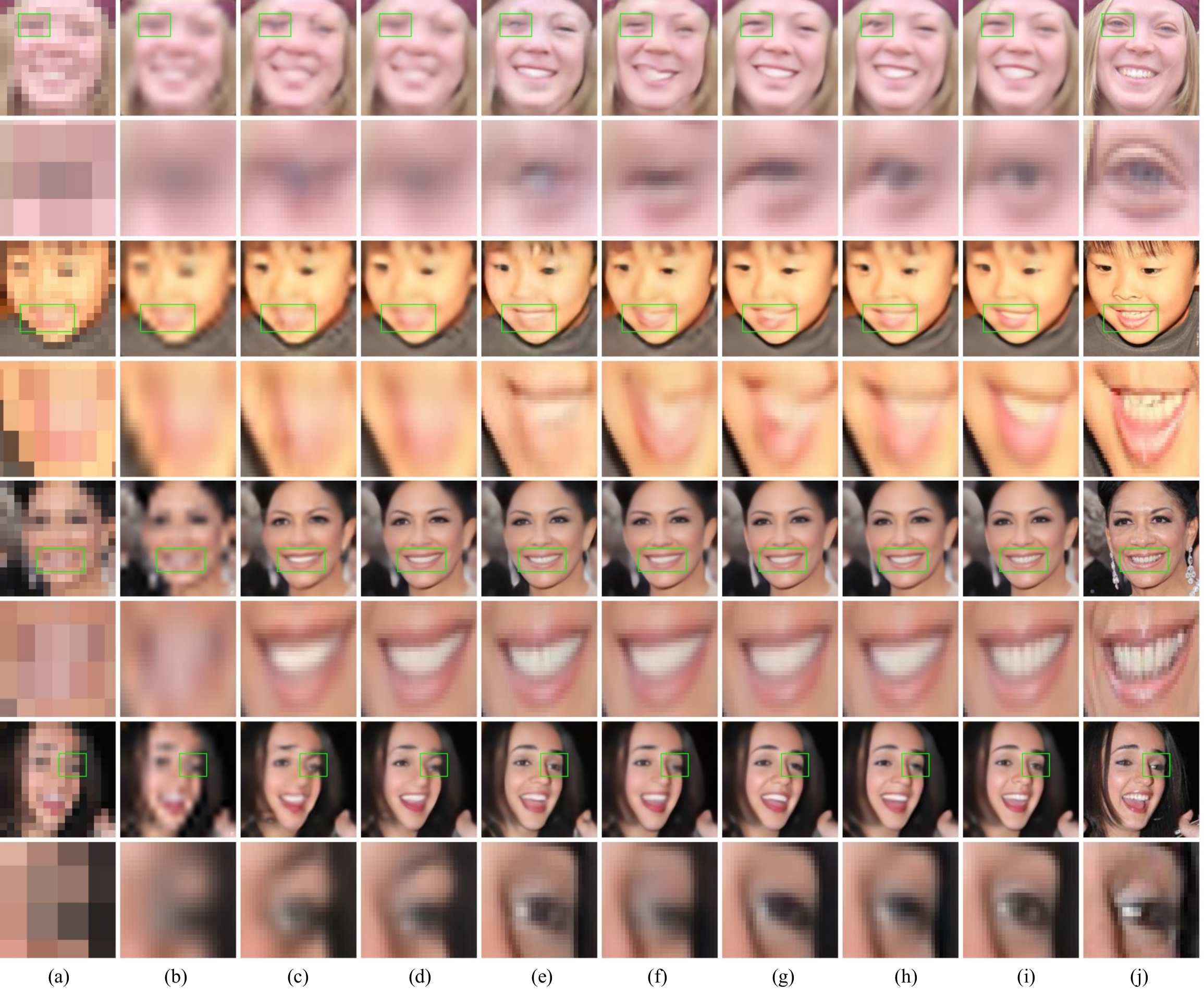}
	\caption{Visual quality comparison of state-of-the-art methods for several side-face examples selected from Helen~\cite{helen} (the first four rows) and CelebA~\cite{celeba} (the last four rows) datasets by the scale of $\times$8. Please zoom in to view the differences. (a): LR; (b): Bicubic; (c): URDGN~\cite{URDGN}; (d): WSRNet~\cite{WaveletSRnet}; (e): Super-FAN~\cite{super-fan}; (f): FSRNet~\cite{FSRNet}; (g): DIC~\cite{DIC}; (h): SISN \cite{SISN}; (i): Ours; (j): HR.}
	\label{fig:upscale_x8_celeba}
\end{figure*}
 
\begin{figure}
	\centering
    \includegraphics[width=\linewidth]{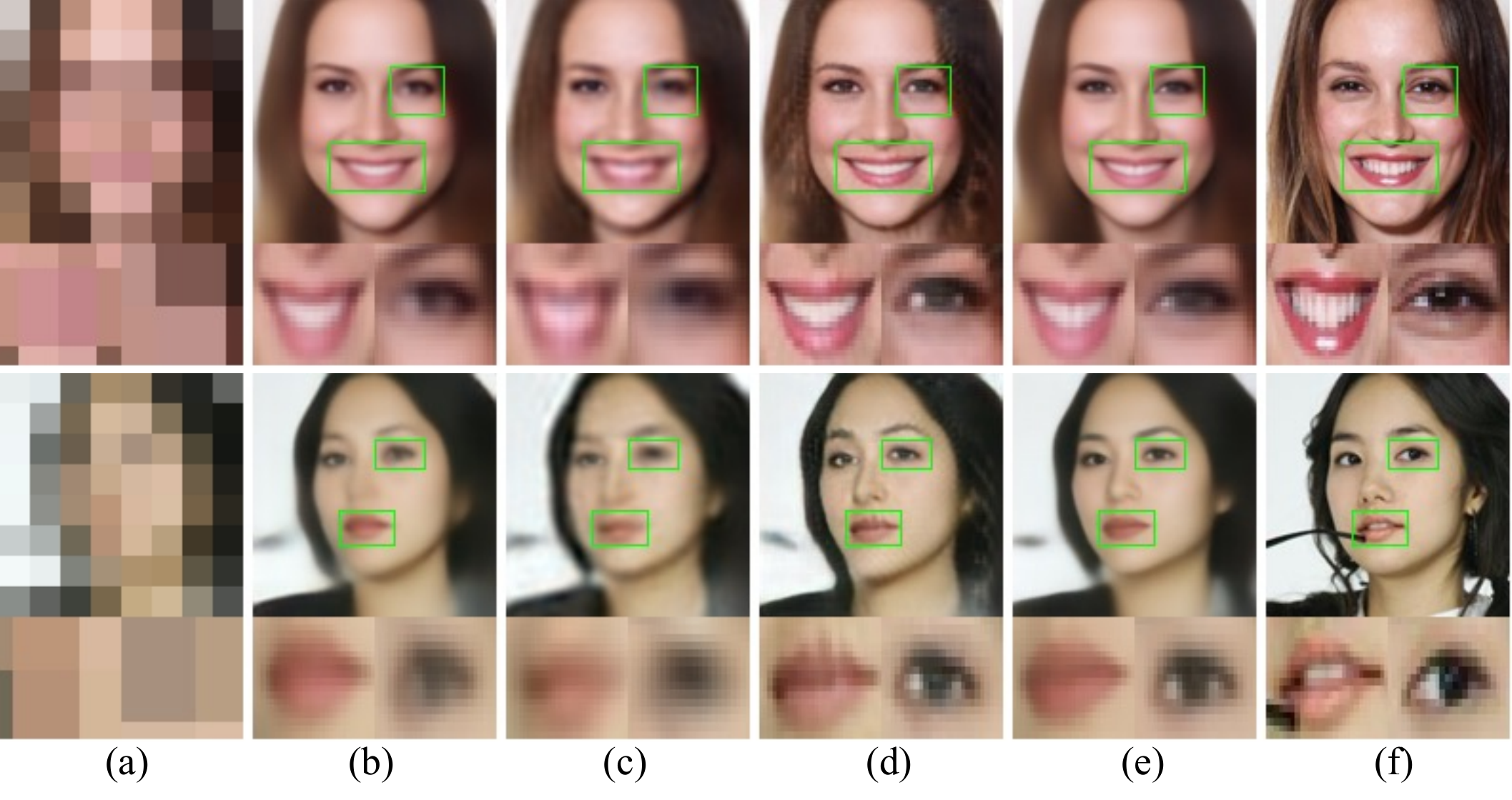}
	\caption{Visual quality comparison of state-of-the-art methods for several face examples selected from CelebA~\cite{celeba} dataset by the scale of $\times$16. Please zoom in to view the differences. (a): LR; (b): DIC~\cite{DIC}; (c): SISN~\cite{SISN}; (d): PSFRGAN \cite{pfsrgan}; (e): Ours; (f): HR.}
	\label{fig:upscale_x16}
\end{figure}

\subsection{Comparisons with State-of-the-Art}
\subsubsection{Qualitative and Quantitative Comparisons on CelebA \cite{celeba} and Helen \cite{helen}} 

      \begin{table*}[t]\renewcommand{\arraystretch}{1.2}
  \caption{Quantitative evaluation of various face hallucination methods on CelebA~\cite{celeba} and Helen~\cite{helen}.}

  \begin{tabularx}{\linewidth}{ m{1.9cm}<{\centering} m{1.1cm}<{\centering}|X<{\centering} X<{\centering} | X<{\centering}  X<{\centering} |X<{\centering} X<{\centering} |X<{\centering}X<{\centering} |X<{\centering} X<{\centering}|X<{\centering} X<{\centering} |X<{\centering}X<{\centering}}
    \hline

      \multirow{3}{*}{Method} & \multirow{3}{*}{Venue}& \multicolumn{6}{c}{CelebA \cite{celeba}} &\multicolumn{6}{c}{Helen \cite{helen}} &\multicolumn{2}{c}{\multirow{2}{*}{Average}}\\ \cline{3-14}
         & & \multicolumn{2}{c}{$\times$4} &\multicolumn{2}{c}{$\times$8} & \multicolumn{2}{c}{$\times$16} & \multicolumn{2}{c}{$\times$4} & \multicolumn{2}{c}{$\times$8} & \multicolumn{2}{c}{$\times$16}     \\ \cline{3-16}
        & & PSNR&SSIM &PSNR&SSIM &PSNR&SSIM&PSNR&SSIM& PSNR&SSIM&PSNR&SSIM &PSNR&SSIM \\
     \hline
     
    Bicubic & -& 27.48 & 0.8166&23.68&0.6258&20.34&0.4671&28.22&0.8540&23.88&0.6628&20.65&0.5061&24.04&0.6554 \\
    SRCNN \cite{SRCNN} &TPAMI'15&28.04&0.8369&23.93&0.6348&20.54&0.4672&28.60&0.8678&23.98&0.6670&20.73&0.5088&24.30&0.6638\\ 
    URDGN \cite{URDGN}&ECCV'16 & 30.11&0.8844& 25.62&0.7261&22.29&0.5786&30.29&0.9019&25.23&0.7205&21.68&0.5452&25.87&0.7261  \\ 
    VDSR \cite{VDSR} &CVPR'16& 31.25&0.9056&26.36&0.7605&22.42&0.5942&30.74&0.9014&25.31&0.7266&21.35&0.5417&26.24&0.7383\\
    WSRNet \cite{WaveletSRnet}& ICCV'17& 30.92&0.9081&26.83&0.7873&23.13&0.6343&30.95&0.9163&26.02&0.7731&22.00&0.5763&26.64&0.7659\\
    Super-FAN \cite{super-fan}  & CVPR'18&31.37&0.9054&27.08&0.7841&23.42&0.6515&30.94&0.9099&26.23&0.7656&22.54&0.5991 &26.93&0.7693\\
   FSRNet \cite{FSRNet} & CVPR'18&31.46&0.9084&26.66&0.7714&23.04&0.6293&30.76&0.9101&25.89&0.7605&22.05&0.5820&26.64&0.7603\\ 
    DIC \cite{DIC}& CVPR'20&31.44&0.9091&\underline{27.41}&\underline{0.8022}&\underline{23.47}&0.6573&\underline{31.81}&\underline{0.9269}&\underline{26.69}&\underline{0.7953}&\underline{22.60}&\underline{0.6122} &\underline{27.24}&0.7838\\ 
         SISN \cite{SISN}& MM'21&\underline{31.88}&\underline{0.9157}&27.31&0.7978&23.42&\underline{0.6743}&31.63&0.9245&26.66&0.7920&22.55&0.6037&\underline{27.24}&\underline{0.7847} \\
    Ours &- &\textbf{31.97}&\textbf{0.9170}&\textbf{27.54}&\textbf{0.8072}&\textbf{23.68}&\textbf{0.6784}&\textbf{32.01}&\textbf{0.9292}&\textbf{26.86}&\textbf{0.7984}&\textbf{22.85}&\textbf{0.6346}&\textbf{27.49}&\textbf{0.7917}\\
    \hline

  \end{tabularx}   
  \label{comparison}
  \vspace{-0.3cm}
  \end{table*} 
To illustrate the superiority of our method in terms of quantitative results, we choose several state-of-the-art methods for comparisons and conduct experiments on CelebA \cite{celeba} and Helen \cite{helen} with multiple upscale factors ($\times$4, $\times$8 and $\times$16). The comparison methods include two representative image super-resolution methods SRCNN~\cite{SRCNN} and VDSR \cite{VDSR}, three general FSR methods URDGN~\cite{URDGN}, WSRNet~\cite{WaveletSRnet} and recently proposed SISN \cite{SISN} without using any face-specific prior knowledge, and three recently proposed face-specific prior-guided FSR methods Super-FAN~\cite{super-fan}, FSRNet~\cite{FSRNet} and DIC~\cite{DIC}. We also introduce the Bicubic interpolation as a baseline comparison method. For a fair comparison, we retrain all the models on our datasets. The quantitative evaluation in terms of PSNR and SSIM metrics are depicted in Table \ref{comparison}, and the visual results of different methods on $\times$4, $\times$8 and $\times$16 are presented in Fig. \ref{fig:upscale_x4}, Fig. \ref{fig:upscale_x8_celeba} and Fig. \ref{fig:upscale_x16}, respectively. In these three figures, we list the LR and ground truth HR in the first and last columns, respectively, and the middle columns are the results of different methods. Here, we present the detailed comparison results. 

\begin{figure}
	\centering
    \includegraphics[width=\linewidth]{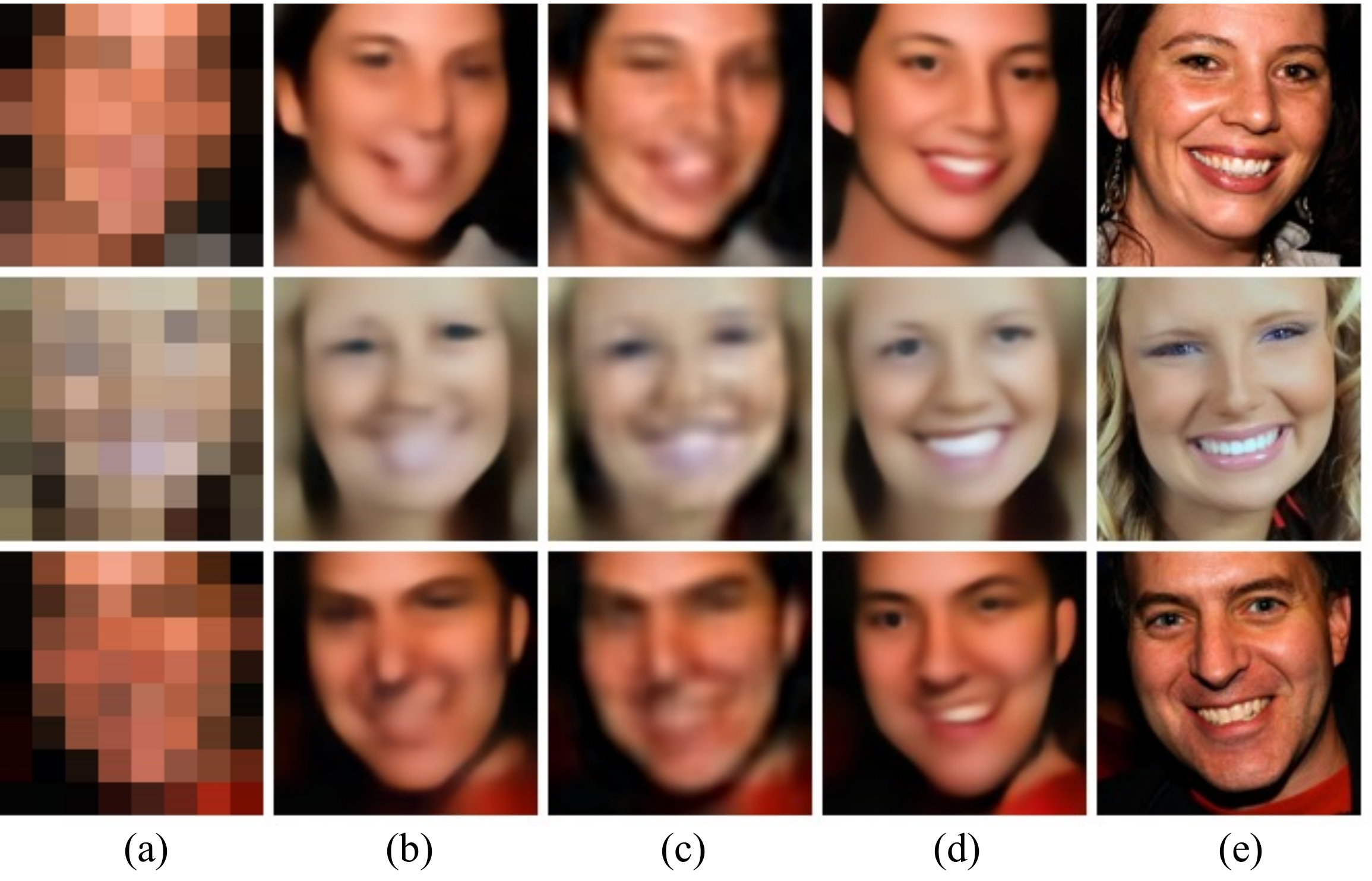}
	\caption{Visual quality comparison of state-of-the-art methods for several side-face examples selected from Helen~\cite{helen} dataset by the scale of $\times$16. Please zoom in to view the differences. (a): LR; (b): DIC~\cite{DIC}; (c): SISN \cite{SISN}; (d): Ours; (e): HR.}
	\label{fig:upscale_x16_helen}
\end{figure}

 SRCNN \cite{SRCNN} is the first deep learning-based image super-resolution method, which is too shallow to recover face images well. The quantitative results of SRCNN are only better than these of Bicubic. URDGN \cite{URDGN} is a generative adversarial network-based FSR method, which pays more attention to recover realistic face images. Due to the property of the generative adversarial network, URDGN is inferior in quantitative metrics and always generates face images with unreal artifacts and missing facial details in visual quality, as shown in Fig. \ref{fig:upscale_x8_celeba}. VDSR \cite{VDSR} is more powerful than SRCNN with a much deeper network. However, it fails to recover face images well. WSRNet \cite{WaveletSRnet} is a wavelet-based FSR method that super-resolves face images in the wavelet domain instead of image domain. Since the wavelet domain can depict the contextual information of the images, WSRNet can recover some high-frequency details. However, WSRNet ignores the utilization of face-specific prior knowledge, resulting in the loss of important facial details. SISN \cite{SISN} builds internal-feature split attention to capture inter-feature information for recovering facial semantic texture. As shown in Fig.~\ref{fig:upscale_x8_celeba}, SISN can recover global facial structure well but cannot reconstruct local facial details, especially on facial components, such as the mouth and eyes. This phenomenon is due to the fact that SISN does not exploit the function of the facial prior. Super-FAN \cite{super-fan} measures the distance between the heatmaps of SR and HR to maintain the structural consistency, named heatmap loss. However, the heatmap loss is only used in the training phase, and face-specific prior knowledge is not used in the testing phase, leading to unfavorable faces, as shown in Fig.~\ref{fig:upscale_x8_celeba}. As shown in Fig.~\ref{fig:upscale_x8_celeba}, FSRNet generates clear faces for $\times$8 FSR, but there are still many artifacts existing, especially on the location of eyes and mouth. When applied for $\times$16 FSR on CelebA~\cite{celeba}, the faces generated by FSRNet are too smooth in Fig.~\ref{fig:upscale_x16}. However, as shown in Fig.~\ref{fig:upscale_x16_helen}, faces recovered by FSRNet have too many artifacts in the experiment of Helen~\cite{helen} with an upscale factor $\times$16. We assume that it is because FSRNet estimates facial prior information from the intermediate results, and the intermediate results are not accurate especially when the upscale factor is large. Therefore, this will lead to accurate prior information derived from the intermediate results, thus inducing artificial results. Similar to FSRNet \cite{FSRNet}, the faces recovered by DIC are blurred and many high-frequency details are lost in $\times$16 FSR on CelebA~\cite{celeba}. For challenging $\times$16 FSR on Helen~\cite{helen}, DIC fails to recover clear facial components.

\begin{figure}
	\centering
    \includegraphics[width=\linewidth]{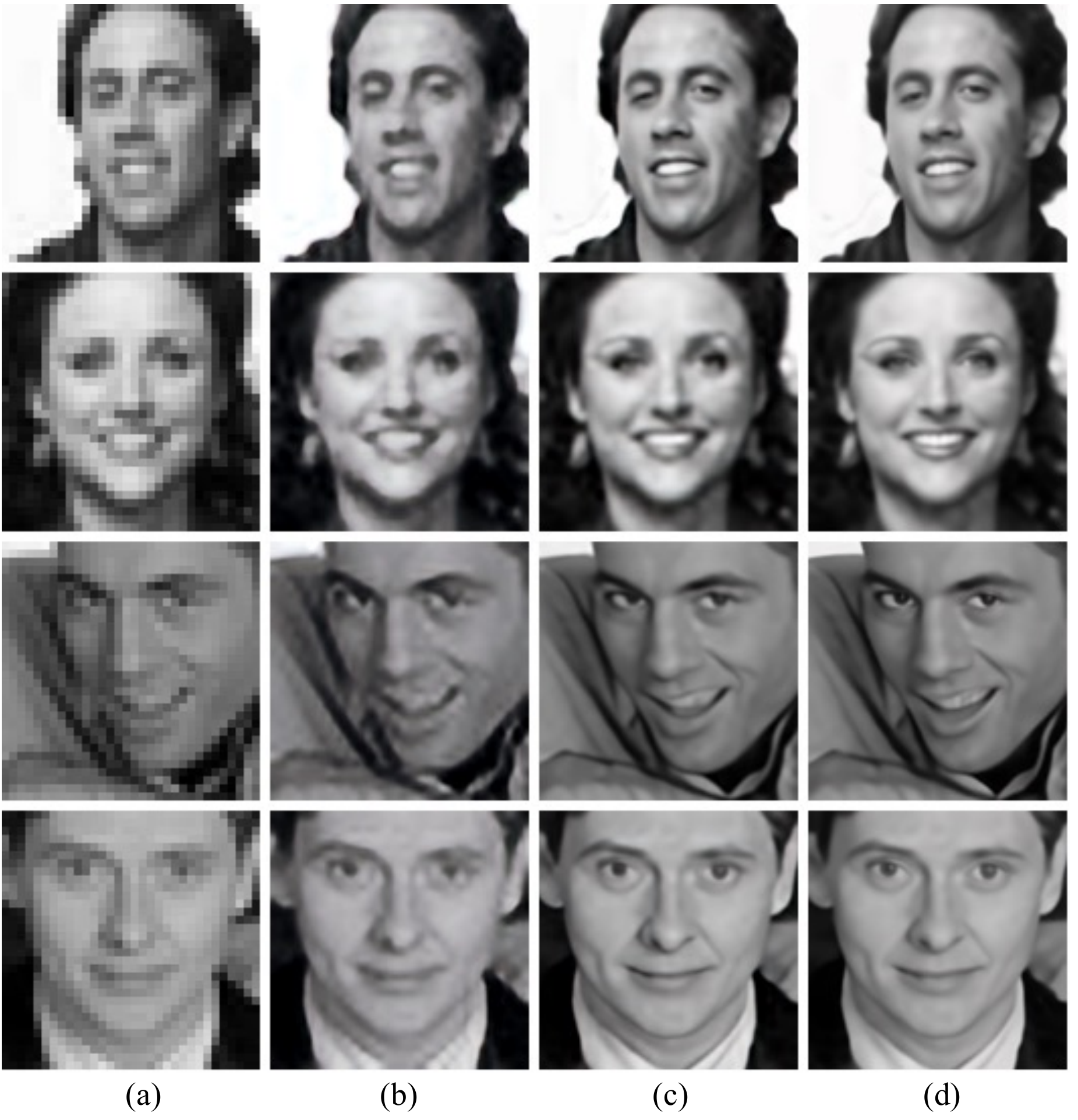}
	\caption{Visual quality comparison of state-of-the-art methods on the real-world face images. (a): Real LR face images; (b) FSRNet \cite{FSRNet}; (c): SISN \cite{SISN}; (d): Ours; Please zoom in to view the differences.}
	\label{fig:real}
\end{figure}

\textbf{Our method} In this paper, we propose a two-stage framework, which first estimates the facial parsing map and then recovers faces under the guidance of the parsing map. Our proposed framework consists of two subnetworks, \textit{i.e.}, ParsingNet and FishFSRNet. The ParsingNet attempts to estimate a facial parsing map directly from LR. This pattern can avoid the bad influence caused by the wrong intermediate results. Then, FishFSRNet enhances the quality of LR with the guidance of the parsing map. Based on the previous FishFSRNet, we further develop a multi-scale refine block to reserve previous multi-scale features and refine the current features. In addition, we build a parsing map attention fusion block which can not only explore inter-channel and inter-spatial relationships, but also make the best use of face-specific prior knowledge. Towards $\times$4 FSR, our methods can repair a clear face image with sharp and realistic details on facial components. As shown in Fig.~\ref{fig:upscale_x8_celeba}, our method can recover rich facial details and holistic facial structure. Although $\times$16 FSR is difficult, our method can still recover visually pleasing face images on CelebA~\cite{celeba}, as shown in Fig.~\ref{fig:upscale_x16}. Compared to FSRNet~\cite{FSRNet}, DIC~\cite{DIC} and SISN \cite{SISN} that fail to recover facial components, our method can recover much clearer facial outline and components even in Helen~\cite{helen} with upscale factor $\times$16, as shown in Fig.~\ref{fig:upscale_x16_helen}. In terms of quantitative metrics, as indicated in Table~\ref{comparison}, our method performs better than other methods on two datasets and all upscale factors. For example, our method outperforms the second-best methods with a large margin of 0.25 dB on Helen~\cite{helen} with upscale factor $\times$16. Compared to these general FSR methods~\cite{SRCNN,URDGN,WaveletSRnet,SISN}, our method takes face-specific prior knowledge into account and achieves better results. Compared to previous prior-guided FSR methods~\cite{super-fan,FSRNet,DIC,pfsrgan}, our method extracts the parsing map from LR and inserts the parsing map into every proposed PAFB to make full use of the parsing map, resulting in superior performance.

\subsubsection{Qualitative Comparisons on Real-world Images} The above experiments are based on simulated datasets, which are significantly different from real-world LR face images. Thus, we apply our method to some real-world faces with unknown degradation. The results are shown in Fig.~\ref{fig:real}. Since the ground truth is not available, we only present the super-resolution results. For comparison, we present the generated results of FSRNet~\cite{FSRNet} and SISN \cite{SISN} in Fig.~\ref{fig:real}. Faces recovered by FSRNet~\cite{FSRNet} have artificial rings and are smooth and blurred. The faces generated by SISN~\cite{SISN} are visually pleasing, but have blurred details on the facial components. In contrast to these methods, our method can recover visually pleasing faces and generate clear and realistic high-frequency details and textures, especially on facial components.

\begin{figure}[t]
	\centering
    \includegraphics[width=\linewidth]{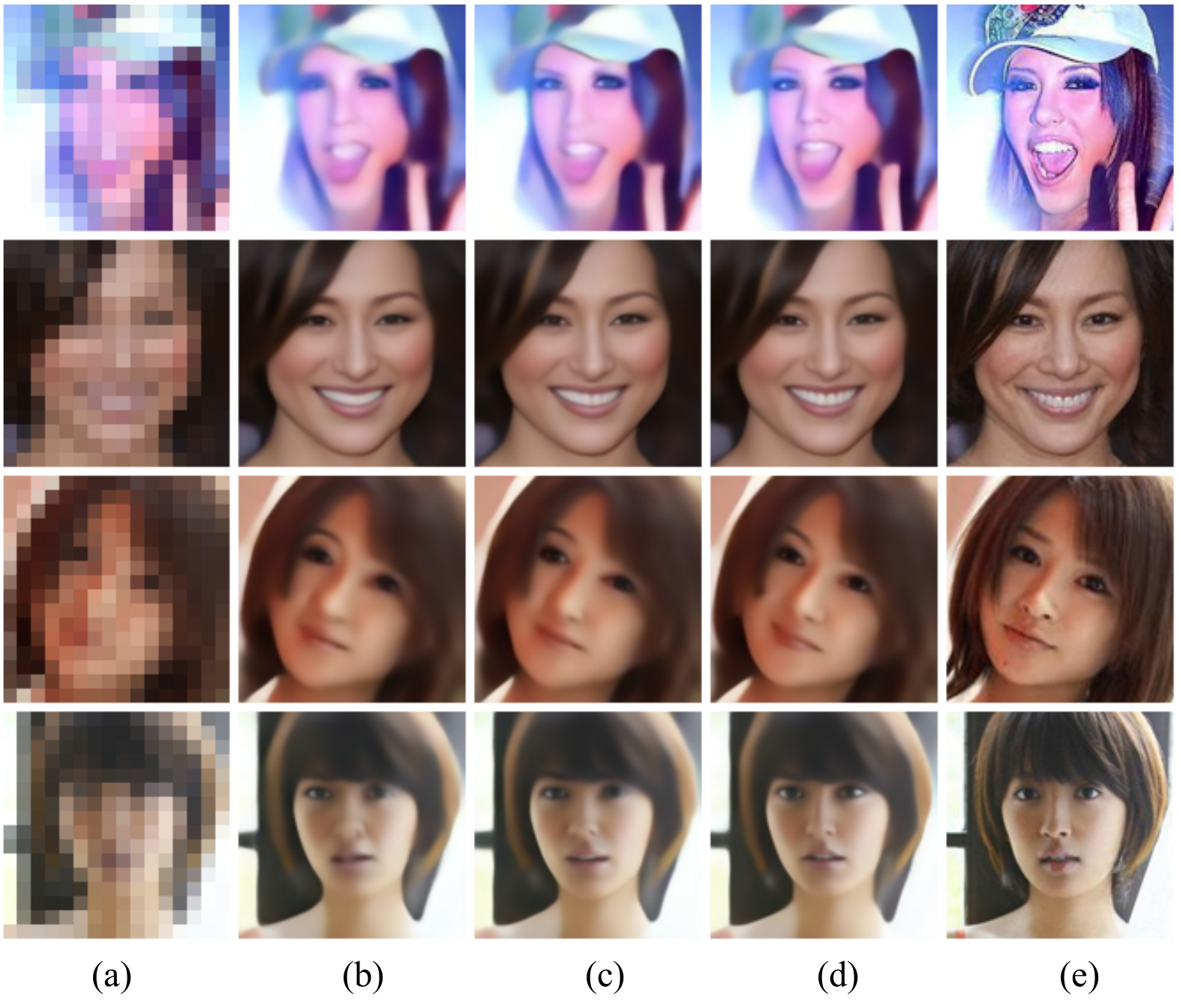}
	\caption{Visual comparison results of different models on CelebA~\cite{celeba} by $\times$8. (a): LR; (b): The super-resolved results of Model 1; (c): The super-resolved result of Model 7; (d): The super-resolved results of Model 8; (e): HR.}
	\label{fig:ablation}
\end{figure}
\begin{table}\renewcommand{\arraystretch}{1.2}
  \caption{Cosine similarity of HR and SR face images recovered by different methods on CelebA \cite{celeba} and Helen \cite{helen} datasets in face recognition.}
    \centering
\begin{tabularx}{\linewidth} { m{1.5cm}<{\centering} X<{\centering}X<{\centering} X<{\centering} X<{\centering}  X<{\centering} X<{\centering} }
    \hline
        \multirow{2}{*}{Datasets}       &\multicolumn{3}{c}{CelebA \cite{celeba}} & \multicolumn{3}{c}{Helen \cite{helen}} \\ \cline{2-7}
         &$\times$4& $\times$8&$\times$16& $\times$4&$\times$8& $\times$16\\
         \hline
         Bicubic& 0.7979& 0.3851& 0.3057& 0.8982& 0.5039& 0.3363\\
         FSRNet \cite{FSRNet}&0.9066& 0.7171& 0.5701 &0.9313& 0.7453& 0.5461\\
        DIC \cite{DIC}& 0.9074& \underline{0.7728}& 0.4975&0.9462& \underline{0.7865}& \textbf{0.5886}\\
         SISN \cite{SISN}&\underline{0.9204}& 0.7709& \underline{0.5919}&\underline{0.9488}& 0.7853& 0.5177\\
         Ours &\textbf{0.9206}&\textbf{0.7780}& \textbf{0.6358} & \textbf{0.9527}& \textbf{0.7960}& \underline{0.5692}\\
         \hline
    \end{tabularx}
    \label{cosine_fr}
\end{table}

\subsubsection{Comparison on Face Recognition} To further verify the superiority of the proposed method, we compare the performance of different FSR methods on face recognition. To be specific, we first obtain the representative vector of the SR and HR face images by pretrained DeepFace \cite{deepface}, and then calculate the cosine similarity of these vectors. The cosine similarity is used to evaluate the face recognition performance of different methods. The results are presented in Table \ref{cosine_fr}. As shown in Table \ref{cosine_fr}, our method achieves the highest cosine similarity in most cases except on $\times$16 Helen \cite{helen}. On $\times$16 Helen, DIC performs best and our method achieves second-best performance. In general, our method can not only improve FSR performance, but also boost face recognition.

\subsection{Ablation Study}
In this section, we analyze and verify the effectiveness of our proposed MSRB and PAFB. To have a clear comparison, we first replace our proposed MSRB and PAFB with concatenation and ResBlock, respectively, and we name this model \textbf{Model 1}. Based on the Model 1, we further conduct a set of experiments and present the quantitative results in Table~\ref{ablation}.

\begin{table}\renewcommand{\arraystretch}{1.2}
  \caption{Effectiveness study of different parts of the proposed network on CelebA with scale $\times$8. AB and PMB denote the attention branch and parsing map branch respectively. SA and CA are spatial attention and channel attention respectively.}
    \centering

    \centering
\begin{tabularx}{0.9\linewidth}{cccccccc}
    \hline
                \multirow{2}{*}{Models} & \multicolumn{2}{c}{PAFB-AB} &\multirow{2}{*}{PAFB-PMB}& \multirow{2}{*}{MSRB} &\multirow{2}{*}{PSNR}&\multirow{2}{*}{SSIM}\\ \cline{2-3}
         & SA &CA  & & & & & \\
         \hline
         1& & && &27.21&0.7951\\
         2&&& & \checkmark &\underline{27.44} &0.8030\\
         3& \checkmark  & \checkmark && &27.31&0.7990\\
         4 & &   & \checkmark&  &27.30 &0.7988\\
         5 &\checkmark &  &\checkmark& &27.35 &0.8013\\
         6 & & \checkmark &\checkmark& & 27.41&0.8028\\
         7 &\checkmark & \checkmark &\checkmark& &\underline{27.44}&\underline{0.8036}\\
         8& \checkmark& \checkmark&\checkmark& \checkmark&\textbf{27.54}&\textbf{0.8072}\\
         \hline
    \end{tabularx}
    \label{ablation}
\end{table}

\begin{figure}[t]
	\centering
    \includegraphics[width=0.95\linewidth]{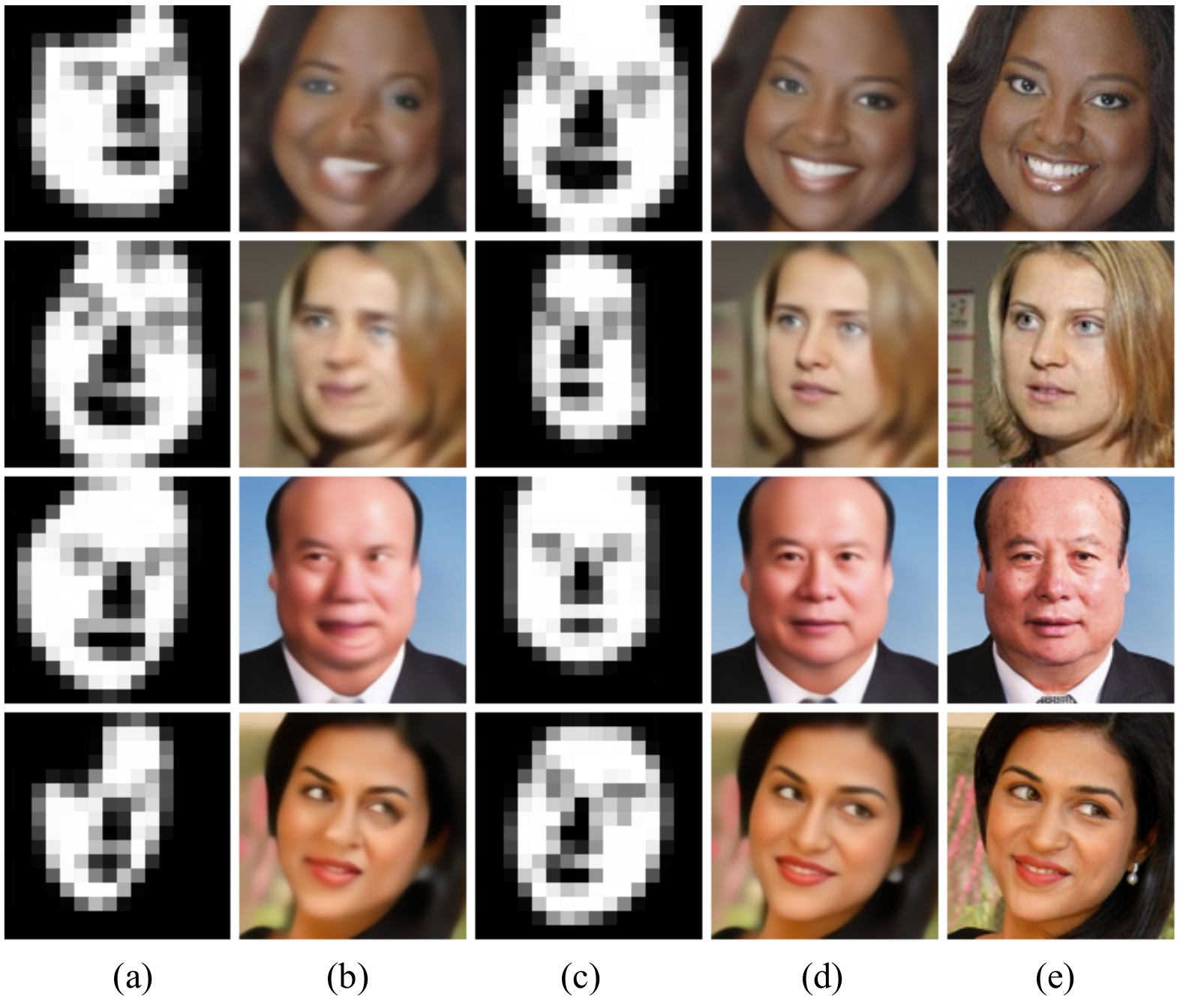}
	\caption{Visual comparison results of different parsing maps on CelebA~\cite{celeba} by $\times$8. (a): Wrong parsing map; (b): The super-resolved result with the wrong parsing map; (c): Right parsing map; (d): The super-resolved result with the right parsing map; (e): HR.}
	\label{fig:ablation_parsing}
\end{figure}
\begin{figure}[t]
	\centering
    \includegraphics[width=\linewidth]{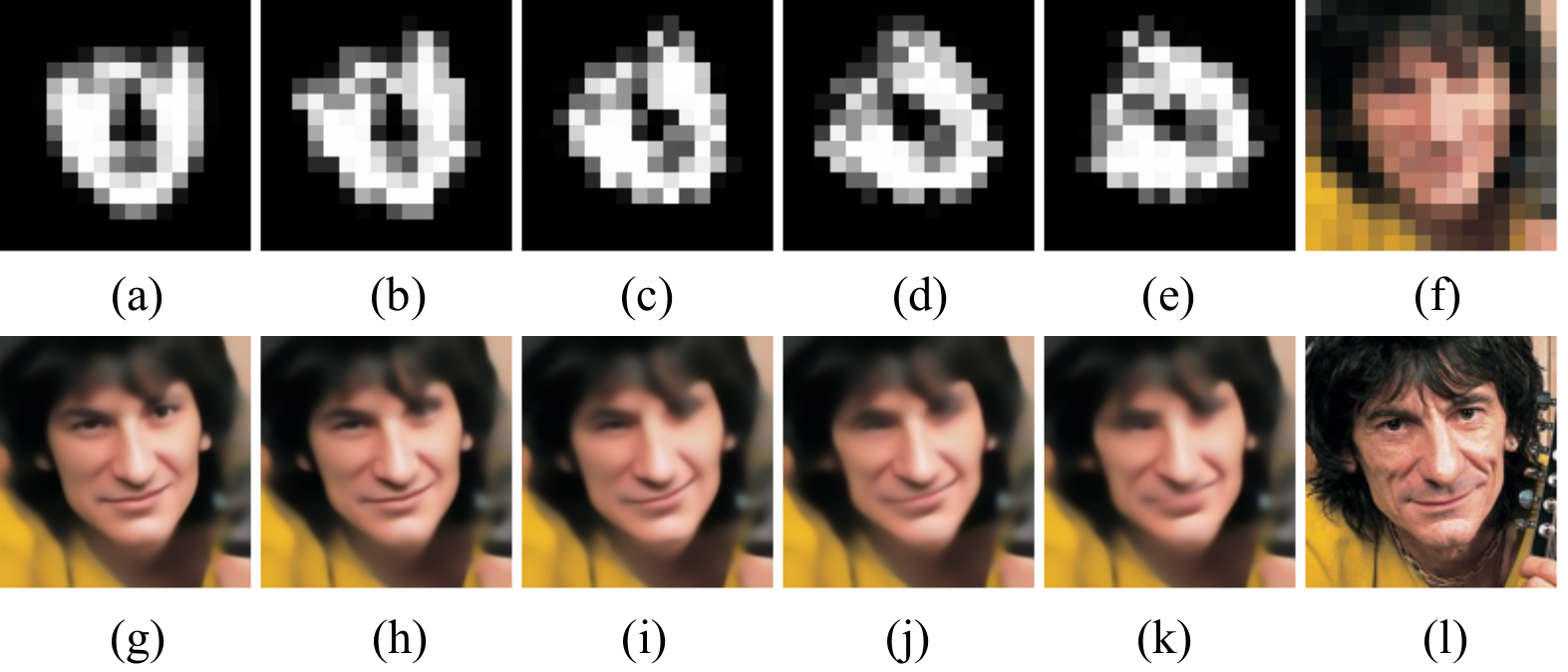}
	\caption{Visual comparison results of turbulent parsing maps on CelebA~\cite{celeba} by $\times$8. (a): Right parsing map; (b): The parsing map rotated by 15 degrees; (c): The parsing map rotated by 30 degrees; (d): The parsing map rotated by 45 degrees; (e): The parsing map rotated by 60 degrees; (f): LR; (g): The super-resolved result with right parsing map; (h): The super-resolved result with parsing map rotated by 15 degrees; (i): The super-resolved result with the parsing map rotated by 30 degrees; (j): The super-resolved result with the parsing map rotated by 45 degrees; (k): The super-resolved result with the parsing map rotated by 60 degrees; (l): HR.}
	\label{fig:ablation_rotate}
\end{figure}
\begin{figure}[t]
	\centering
    \includegraphics[width=\linewidth]{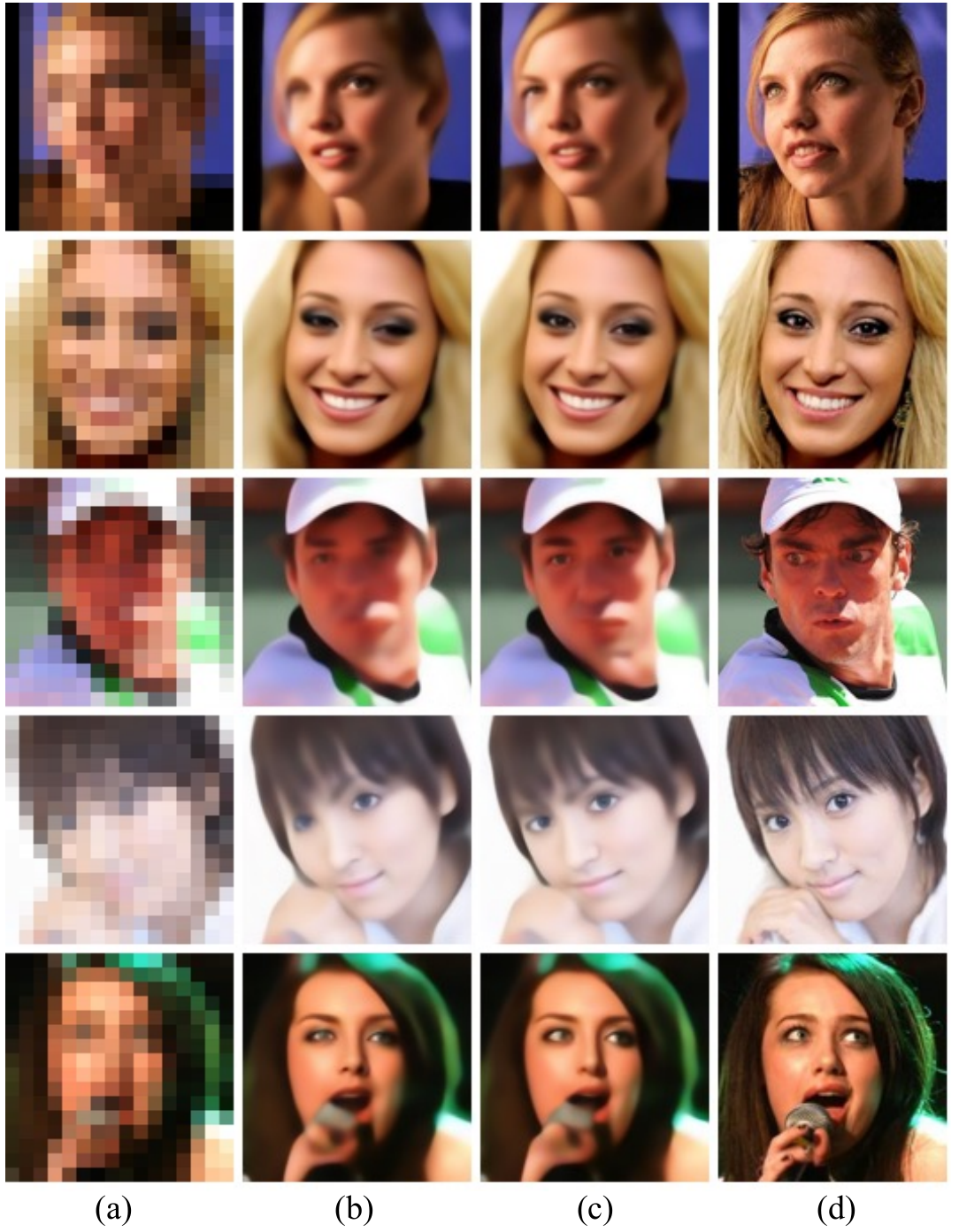}
	\caption{Visual quality comparison of the model with the parsing map and the model without the parsing map. Please zoom in to view the differences. (a): LR; (b): The results of the model without the parsing map; (c): The results of the model with the parsing map; (d): HR.}
  \vspace{-0.1in}
	\label{fig:with_parsing}
\end{figure}

\textbf{The Effectiveness of MSRB:} Compared with our previous work~\cite{our}, we develop a MSRB to boost the feature representation ability. To verify the effectiveness of it, we directly replace the concatenation in Model 1 with our proposed MSRB, and name it \textbf{Model 2}. Then, we compare the performance of Model 1 and Model 2. From the Table~\ref{ablation}, it is obvious that the performance of Model 2 is much better than that of Model 1, which proves that our MSRB can boost the FSR effectively. From the prospective of visual quality, we present the results of Model 1 and Model 2 in Fig.~\ref{fig:ablation}. The visual FSR results of Model 2 is globally clearer than the ones of Model 1. However, our proposed MSRB is not tailored for faces and leaves out face-specific prior knowledge. Thus, the super-resolved results on facial components is not sharp enough.

\begin{table}[t]\renewcommand{\arraystretch}{1.2}
\centering
  \caption{Face Parsing Accuracy on CelebA \cite{celeba} and Helen \cite{helen}.}
\begin{tabularx}{0.9\linewidth} {X<{\centering}X<{\centering}X<{\centering} X<{\centering} X<{\centering}  X<{\centering} X<{\centering} }
    \hline
             \multirow{2}{*}{Dataset} &\multicolumn{3}{c}{CelebA \cite{celeba}} & \multicolumn{3}{c}{Helen \cite{helen}} \\ \cline{2-7}
         &$\times$4& $\times$8&$\times$16&$\times$4&$\times$8&$\times$16\\
         \hline
         Accuracy& 0.9763& 0.9671&0.9564&0.9520&0.9228&0.8672\\

         \hline
    \end{tabularx}
    \label{parsing_accuracy}
\end{table}

\textbf{The Effectiveness of PAFB:} In Section~\ref{des-pafb}, we have introduced our PAFB which contains two branches (\textit{i.e.}, an attention branch and a parsing map branch). Here, we conduct another series of experiments to validate the effectiveness of our PAFB. First, we verify the effectiveness of two branches respectively. We only reserve the attention branch and remove the parsing map branch (it should be noted that the concatenation followed by a convolution layer at the end of PAFB is also discarded), and this model is called \textbf{Model 3}. At this time, Model 3 only explores the inter-channel and inter-spatial relationship without considering face-specific prior knowledge. In contrast to Model 3, we remove the attention branch and only reverse the parsing map branch in PAFB, which is \textbf{Model 4}. As shown in Table~\ref{ablation}, with the parsing map providing facial structure information, Model 4 can improve the performance of FSR, verifying the effectiveness of face-specific knowledge. In addition, we conduct experiments to analyze the function of different attention mechanisms in FSR when the parsing map branch is used. Specially, we remove the channel attention (CA) or spatial attention (SA) from the PAFB, and introduce the edited PAFB to Model 1, generating \textbf{Model 5} and \textbf{Model 6} respectively. From Table \ref{ablation}, we can find that both SA and CA can promote FSR under the condition that the parsing map is explored. We analyze that facial parsing map provides facial structure information which is a semantic-level guidance, while the attention mechanism aims to explore relationship along spatial and channel dimensions, which can be viewed as pixel-level information. Thus, they are complementary, and the collaboration between them can enhance the representation ability of the network, promoting FSR. Finally, we add MSRB to Model 7, generating \textbf{Model 8}. Clearly, Model 8 achieves the best performance in terms of PSNR and SSIM. In terms of visual quality, the results of Model 1, Model 2 and Model 8 are all presented in Fig.~\ref{fig:ablation}. Due to the consideration of face-specific prior knowledge, Model 8 not only recovers more visually pleasing face images than Model 1 and Model 2 globally, but also reconstructs much sharper details on the facial components such as eyes and mouths. In summary, the effectiveness of every proposed component is verified by the above experiments.

\textbf{The Effectiveness of Parsing Map:} In addition, we also compare the visual results of different parsing maps to present the influence of the parsing maps. As shown in Fig.~\ref{fig:ablation_parsing}, given an LR face, we feed the LR with a right parsing map $P$ shown in Fig.~\ref{fig:ablation_parsing}(c), and then our model generates $I_{\text{SR}}$ (shown in Fig.~\ref{fig:ablation_parsing}(d)) which is very close to the ground truth. When we give the model a wrong or mismatched parsing map $P$ shown in Fig.~\ref{fig:ablation_parsing}(a), we obtain the $I_{\text{SR}^{1}}$ (shown in Fig.~\ref{fig:ablation_parsing}(b)) which has many artifacts and distorted facial components. Clearly, a wrong parsing map would lead to distorted face images, while a correct parsing map can boost the FSR. Except that, the results also verify that our model makes full use of face-specific prior knowledge. Considering the huge differences between the wrong and right parsing maps, we give some disturbance to the right parsing and conduct another experiment. To be specific, we also rotate the right parsing map by 15, 30, 45, 60 degrees, and then obtain the super-resolved results with these disturbed parsing maps. As shown in Fig.~\ref{fig:ablation_rotate}, the greater the disturbance, the worse the visual quality. 

In addition, we present the face parsing accuracy of our ParsingNet in Table \ref{parsing_accuracy} and present a visual comparison between our final model and our model without the parsing map in Fig. \ref{fig:with_parsing} to further verify the effectiveness of the parsing map. To be specific, we remove the parsing map in the parsing map branch and keep parameter similar, and train the revised model. Then, we show the visual comparison between these two models in Fig. \ref{fig:with_parsing}. We can see that our model with the parsing map can recover more visually pleasing face images than the model without the parsing map. Especially in the regions of facial components, the face images hallucinated by our method have much clearer texture and details. By these experiments, we can conclude that the facial parsing map plays an important role in FSR.                                                        

\section{Conclusion}
This paper proposes a parsing map attention fusion network for face hallucination. In particular, it directly estimates a facial parsing map from LR, and then recovers the final face images with the extracted parsing map. Based on this, we further develop a novel multi-scale refine block to reserve the previous features of different resolutions and refine the current features, which can significantly improve the performance. Except that, we build a parsing map attention fusion block which can not only explore the relationship among channel and spatial dimensions but also exploit the contribution of face-specific prior (parsing map). Finally, the paper conducts massive comparison experiments to verify the effectiveness of every proposed component, and the proposed method can achieve favorable performance against existing methods.

% or


% \appendices

% \section{}

% \section*{Acknowledgment}


\ifCLASSOPTIONcaptionsoff
  \newpage
\fi


\bibliographystyle{IEEEtran}
\bibliography{Mendeley}

% \normalem
% \bibliographystyle{IEEEtran}
% \bibliography{Mendeley.bib}

\end{document}